\documentclass[sigconf]{acmart}
\copyrightyear{2025}
\acmYear{2025}
\setcopyright{acmlicensed}\acmConference[KDD '25]{Proceedings of the 31st ACM SIGKDD Conference on Knowledge Discovery and Data Mining V.2}{August 3--7, 2025}{Toronto, ON, Canada}
\acmBooktitle{Proceedings of the 31st ACM SIGKDD Conference on Knowledge Discovery and Data Mining V.2 (KDD '25), August 3--7, 2025, Toronto, ON, Canada}
\acmDOI{10.1145/3711896.3736961}
\acmISBN{979-8-4007-1454-2/2025/08}
\AtBeginDocument{%
  \providecommand\BibTeX{{%
    \normalfont B\kern-0.5em{\scshape i\kern-0.25em b}\kern-0.8em\TeX}}}

\usepackage{amsmath,amsfonts}

\usepackage{amssymb}
\usepackage{algorithmic}
\usepackage{graphicx}
\usepackage{textcomp}
\usepackage{xcolor}
\usepackage{multirow} %合并行
\usepackage{booktabs} 
\usepackage{bigstrut}
\usepackage{tabu}
\usepackage{tabularx} % for 'tabularx' env. and 'X' col. type
\usepackage{ragged2e} % for \RaggedRight macro
\usepackage{subfigure}
\usepackage{enumitem}
\usepackage{rotating}
\setlist[itemize]{leftmargin=10pt}
\usepackage{algorithmic}
\usepackage[linesnumbered,ruled,vlined]{algorithm2e}
\def\BibTeX{{\rm B\kern-.05em{\sc i\kern-.025em b}\kern-.08em
    T\kern-.1667em\lower.7ex\hbox{E}\kern-.125emX}}

\begin{document}

%%
%% The "title" command has an optional parameter,
%% allowing the author to define a "short title" to be used in page headers.
\title{Fine-Grained Traffic Inference from Road to Lane via Spatio-Temporal Graph Node Generation}

%%
%% The "author" command and its associated commands are used to define
%% the authors and their affiliations.
%% Of note is the shared affiliation of the first two authors, and the
%% "authornote" and "authornotemark" commands
%% used to denote shared contribution to the research.
\author{Shuhao Li}
\affiliation{%
  \institution{ Fudan University \\ Shanghai Key Laboratory of Data Science}
  \city{Shanghai}
  \country{China}}
\email{shli23@m.fudan.edu.cn}

\author{Weidong Yang}
\authornote{Corresponding author.}
\affiliation{%
  \institution{Fudan University}
  \city{Shanghai}
  \country{China}}
\affiliation{%
  \institution{Zhuhai Fudan Innovation Research Institute }
  \city{Zhuhai}
  \country{China}}
\email{wdyang@fudan.edu.cn}

\author{Yue Cui}
\affiliation{%
  \institution{The Hong Kong University of Science and Technology}
  \city{Hong Kong SAR}
  \country{China}}
\email{ycuias@cse.ust.hk}

\author{Xiaoxing Liu}
\affiliation{%
  \institution{Guangzhou University \\ GZHU-SCHB Intelligent Transportation Joint Lab}
  \city{Guangzhou}
  \country{China}}
\email{xiaoxingliu@e.gzhu.edu.cn}

\author{Lingkai Meng}
\affiliation{%
  \institution{Shanghai Jiao Tong University}
  \city{Shanghai}
  \country{China}}
\email{mlk123@sjtu.edu.cn}

\author{Lipeng Ma}
\affiliation{%
  \institution{ Fudan University \\ Shanghai Key Laboratory of Data Science}
  \city{Shanghai}
  \country{China}}
\email{lpma21@m.fudan.edu.cn}

\author{Fan Zhang}
\affiliation{%
  \institution{ Guangzhou University \\ GZHU-SCHB Intelligent Transportation Joint Lab}
  \city{Guangzhou}
  \country{China}}
\email{zhangf@gzhu.edu.cn}

%%
%% By default, the full list of authors will be used in the page
%% headers. Often, this list is too long, and will overlap
%% other information printed in the page headers. This command allows
%% the author to define a more concise list
%% of authors' names for this purpose.
\renewcommand{\shortauthors}{Shuhao Li et al.}
% \thanks{$\dagger$ Corresponding author.}

%%
%% The abstract is a short summary of the work to be presented in the
%% article.
\begin{abstract}

Fine-grained traffic management and prediction are fundamental to key applications such as autonomous driving, lane change guidance, and traffic signal control. However, obtaining lane-level traffic data has become a critical bottleneck for data-driven models due to limitations in the types and number of sensors and issues with the accuracy of tracking algorithms. To address this, we propose the \underline{\textbf{F}}ine-grained \underline{\textbf{R}}oad \underline{\textbf{T}}raffic \underline{\textbf{I}}nference (\textbf{FRTI}) task, which aims to generate more detailed lane-level traffic information using limited road data, providing a more energy-efficient and cost-effective solution for precise traffic management. This task is abstracted as the first scene of the spatio-temporal graph node generation problem. We designed a two-stage framework—\textbf{RoadDiff}—to solve the FRTI task. This framework leverages the Road-Lane Correlation Autoencoder-Decoder and the Lane Diffusion Module to fully utilize the limited spatio-temporal dependencies and distribution relationships of road data to accurately infer fine-grained lane traffic states. Based on existing research, we designed several baseline models with the potential to solve the FRTI task and conducted extensive experiments on six datasets representing different road conditions to validate the effectiveness of the RoadDiff model in addressing the FRTI task. The relevant datasets and code are available at https://github.com/ShuhaoLii/RoadDiff.
\end{abstract}

%%
%% The code below is generated by the tool at http://dl.acm.org/ccs.cfm.
%% Please copy and paste the code instead of the example below.
%%
%%
%% Keywords. The author(s) should pick words that accurately describe
%% the work being presented. Separate the keywords with commas.
% \keywords{Traffic Prediction, Graph Neural Network, Fine-Grained Prediction, Spatio-Temporal Data}

%%
%% This command processes the author and affiliation and title
%% information and builds the first part of the formatted document.
\maketitle
\section{Introduction}
Lane-level traffic prediction plays a crucial role in the development of smart cities, significantly supporting applications such as dynamic lane allocation, managing peak-period traffic, and configuring traffic signals. In recent years, an increasing number of studies have focused on optimizing lane-level traffic prediction schemes \cite{li2024unifying,lu2020lane,zhou2024maginet}, resulting in a notable improvement in prediction accuracy. However, the development of lane-level traffic prediction still faces challenges, primarily due to the limited number of dedicated cameras and the complexity and data quality issues associated with obtaining lane-level speed and flow information from videos. These factors have led to lane-level traffic prediction lagging behind road-level traffic prediction.

\begin{figure}[t]
\vspace{-0.15cm}
\centerline{\includegraphics[width=0.45\textwidth]{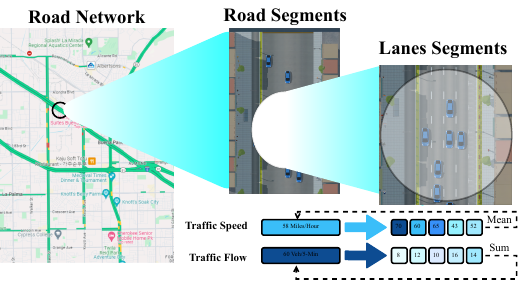}}
\vspace{-0.15cm}
\caption{Examples of inferring lane data from roads and the differences in inferring traffic speed and flow.}
% \vspace{-0.15cm}
\label{fig: intro}
\end{figure}

In addition to specialized lane-level models, existing studies have achieved good lane-level predictions using road-level models \cite{li2024unifying,li2024st}, but they overlook correlations between different granularities and traffic states. As shown in \textbf{Figure~\ref{fig: intro}}, we analyzed the different constraints between road traffic states and lane traffic states, e.g., the mean lane speed equals the road speed, while the flow is the sum of individual lane flows. To efficiently utilize road-level traffic information for fine-grained lane-level downstream traffic tasks and to advance the development of lane-level traffic prediction, we propose a new research task—the fine-grained road traffic inference (FRTI) task. This problem aims to infer detailed lane-level traffic information using road traffic data and the corresponding lane topology, thereby promoting downstream tasks in a more environmentally friendly, cost-effective, and efficient manner.

The FRTI task presents several challenges. \textbf{First, coarse-grained road traffic information contains limited details}, making it difficult to reflect lane-level specifics. Existing road-level data generally provide overall traffic conditions \cite{cui2021metro,bai2020adaptive,cui2023roi} but fail to capture specific flow and speed variations for each lane. Inferring richer lane-level dependencies from limited road-level data while addressing extensive uncertainties poses significant challenges.

\textbf{Different segments of the same road may have varying numbers of lanes}, making it difficult to apply existing grid-based modeling approaches, such as those used in image super-resolution tasks and fine-grained urban flow inference (FUFI) tasks \cite{liang2019urbanfm,li2022fine,qu2022forecasting}. Treating the road network as a grid structure ignores the irregularities of roads and lanes, leading to missing or non-existent lane segments, which affects the accuracy and practicality of the models.

% Different segments of the same road may have varying numbers of lanes, complicating the direct application of methods from the fine-grained urban traffic flow inference (FUFI) problem. Similar to image super-resolution, FUFI solutions often rely on a gridded data structure. However, treating the road network as a gridded structure overlooks the irregularity of roads and lanes, leading to omissions or nonexistent lane segments, affecting model accuracy and practicality.

\textbf{Additionally, roads and lanes differ in speed and flow dependencies and application demands}. As shown in \textbf{Figure \ref{fig: intro}}, using the Santa Ana Freeway in Los Angeles as an example, speed and flow are governed by distinct constraints. Traffic states vary across data sources: loop detectors and cameras provide lane-level flow information, while floating vehicles typically capture speed data. Application requirements also differ; for instance, lane guidance prioritizes the fastest lane, while logistics fleets prefer low-flow lanes to minimize lane changes. Models must address these differences to accurately infer traffic states under varying constraints.

\begin{figure}[t]
\vspace{-0.15cm}
\centerline{\includegraphics[width=0.42\textwidth]{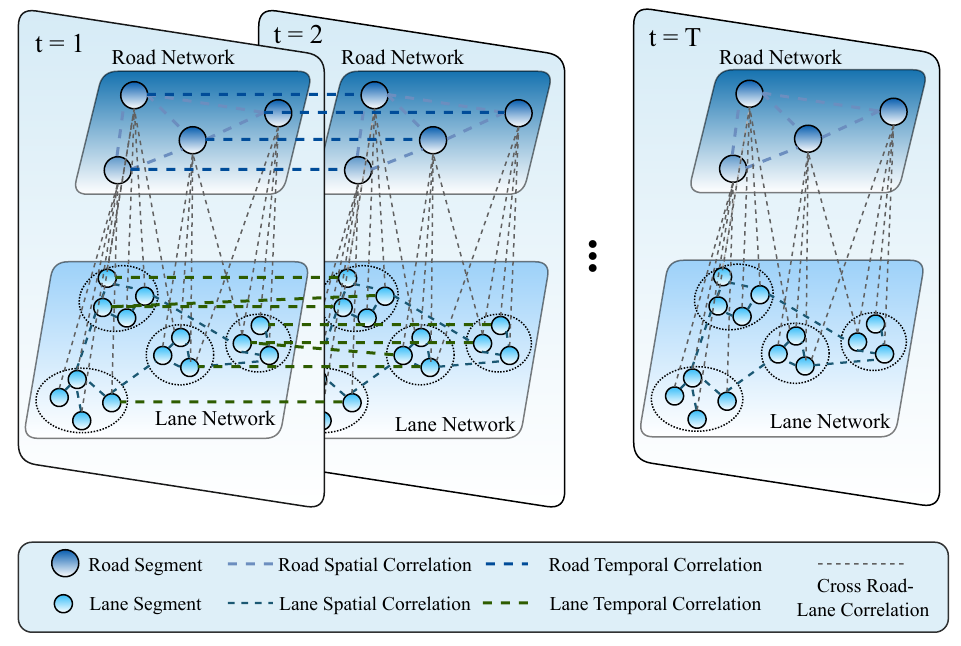}}
\vspace{-0.25cm}
\caption{The spatio-temporal graph node generation problem of generating a lane network from a road network and the complex dependencies faced by the FRTI task.}
\label{fig: intr_node_g}
\end{figure}

In summary, the main challenges of FRTI include modeling and deducing the limited yet complex spatio-temporal relationships between road and lane level data, the fixed yet irregular topology of lane networks, and the different constraints of speed and flow between lanes and roads. To address these issues, as illustrated in \textbf{Figure \ref{fig: intr_node_g}}, we use graph structure to model the road and lane networks simultaneously, abstracting the FRTI task as a spatio-temporal graph node generation problem under complex constraints, which differs from graph reconstruction problem that does not involve an increase in the number of nodes \cite{wen2023diffstg,hu2023towards}. We designed the RoadDiff framework as the solution to this problem. RoadDiff employs a simple yet efficient encoder-decoder architecture to model the spatio-temporal dependencies of road and initial lane information in detail. It then uses a lane diffusion module to refine lane information generation and learn the constraints between different traffic states under varying demands. 

Our main contributions are summarized as follows:
\begin{itemize}
    \item We are the first to propose and formalize the FRTI task, defining it as a spatio-temporal graph generation problem under complex conditions. This is the first node generation problem for spatio-temporal graph data, presenting multiple unique challenges.

    \item We designed a road-lane autoencoder module. In the encoding phase, we comprehensively model the spatio-temporal dependencies of the road network using a simple network architecture. In the decoding phase, we aggregate and restore lane information through the lane network topology and spatio-temporal attention mechanisms, generating the initial lane information.

    \item We developed a lane diffusion module to refine the initial lane information, learning the uncertainties and the constraints between different traffic states, ensuring the accuracy and regulations compliance of the generated fine-grained lane information.

    \item We conducted traffic speed and flow inference experiments on six real datasets. We improved existing models for their adaptability to the FRTI problem, using them as baseline models for comparison. The experimental results show that RoadDiff performs excellently in inferring both types of traffic states.
\end{itemize}
% \vspace{-0.3cm}
\section{Problem Formulation}

In this section, we define the notations and then elaborate on the fine-grained road traffic inference (FRTI) task.

\textbf{Definition 1 (Road Network)}: We define the road network as an undirected graph \( G^R = (V^R, E^R, A^R) \), where each node \( r_i \in V^R \) represents the \( i \)-th road segment, and each edge \( e^r \in E^R \) represents the connections between road segments. \( A^R \in \mathbb{R}^{I \times I} \) denotes the static adjacency matrix of the road network.

\begin{figure}[t]
% \vspace{-0.45cm}
\centerline{\includegraphics[width=0.4\textwidth]{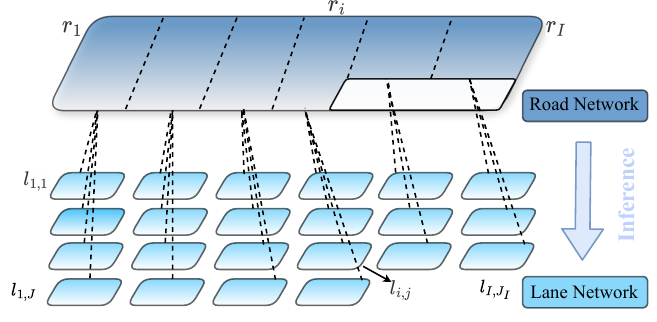}}
\vspace{-0.25cm}
\caption{The correspondence between roads and lanes in the FRTI task.}
%\vspace{-0.45cm}
\label{fig: intro_r}
\end{figure}

\textbf{Definition 2 (Lane Network)}: To model the regular and irregular number of lanes present in real-world road scenarios, as exemplified in \textbf{Figure \ref{fig: intro_r}}, we also define the lane network as an undirected graph \( G^L = (V^L, E^L, A^L) \), where each node \( l_{i,j} \) represents the \( j \)-th lane segment of the \( i \)-th road segment, and each edge \( e^l \in E^L \) represents the adjacency relationship between lanes. We define the adjacency relationships in four directions: up, down, left, and right. \( A^L \in \mathbb{R}^{N \times N} \) denotes the static adjacency matrix of the lane network, where \( N = \sum_{i=0}^I J_i \), and \( J_i \) represents the maximum number of lanes for the corresponding road segment \( r_i \).

Given a time interval \( t \), we use \( x^{r_i}_t \) to represent the average traffic state of road segment \( r_i \) at time $t$, and \( x^{l_{i,j}}_t \) to represent the average traffic state of lane segment \( l_{i,j} \) at time $t$. \( X^R_t = [x_t^{r_1}, x_t^{r_2}, \ldots, x_t^{r_I}] \) represents the traffic state of the entire road network at time \( t \), and \( X^L_t = [x_t^{l_{1,1}}, x_t^{l_{1,2}}, \ldots, x_t^{l_{I, J_I}}] \) represents the traffic state of the entire lane network at time \( t \).

\textbf{Constraint 1 (Traffic Speed Constraint)}: The average traffic speed of the lanes corresponding to road \( r_i \) must equal the traffic speed of the road at time $t$, i.e., \(x_t^{r_i} = \frac{1}{J_i} \sum_{j=1}^{J_i} x_t^{l_{i,j}} \).

\textbf{Constraint 2 (Traffic Flow Constraint)}: The total traffic flow of the lanes corresponding to road \( r_i \) must equal the traffic flow of the road at time $t$, i.e., \( x_t^{r_i} = \sum_{j=1}^{J_i} x_t^{l_{i,j}} \).

\textbf{Problem Definition (FRTI)}: Given the road network traffic state \( X^R \) over a time window of size \( T \), \( X^R = \{X_1^R, X_2^R, \ldots, X_T^R\} \), representing the traffic states (speed or flow) of all nodes in the road network over \( T \) time slices. The task of fine-grained road traffic inference aims to use \( X^R \) and the lane network topology to infer the fine-grained lane traffic state \( X^L \), \( X^L = \{X_1^L, X_2^L, \ldots, X_T^L\} \), in the following form:
\begin{equation}
    f[X^R, G^R, G^L] \rightarrow X^L
\end{equation}
\section{Method}
\begin{figure*}[t]
\vspace{-0.15cm}
	\centering
		\includegraphics[width=0.8\linewidth]{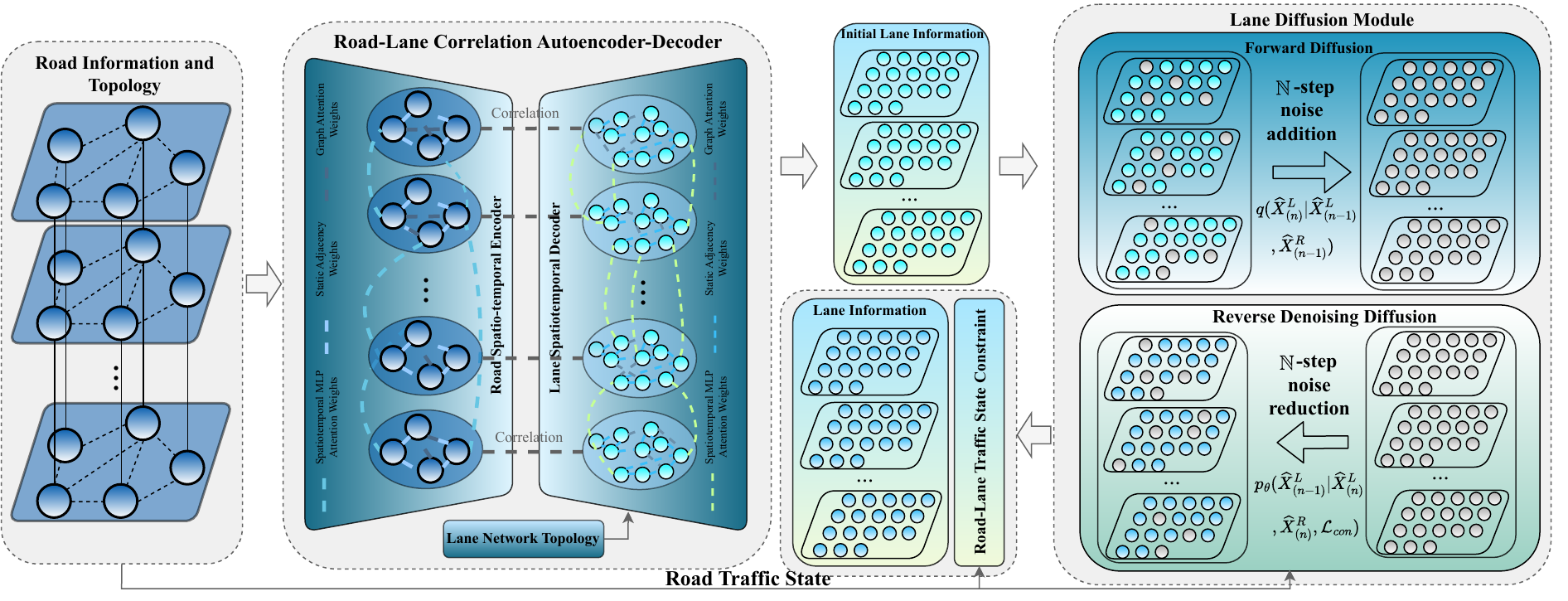}
  \vspace{-0.15cm}
		\caption{Overview of the two-stage RoadDiff framework:  the first stage, with a streamlined spatio-temporal attention design, provides a solid starting point for the diffusion stage, which further refines lane uncertainty and enforces physical constraints.}
		\label{fig: RoadDiff}
  \vspace{-0.25cm}
\end{figure*}
We proposed the RoadDiff framework for fine-grained road traffic inference, considering spatio-temporal dependencies and real-world traffic constraints. As shown in \textbf{Figure \ref{fig: RoadDiff}}, RoadDiff includes two main components and a constraint output module. The Road-Lane Autoencoder-Decoder models spatio-temporal dependencies in the road network. The encoder extracts high-dimensional features, while the decoder reconstructs lane network states from these features, generating lane-level information. Initial lane information may have biases due to the limitations of end-to-end generation \cite{xiao2023upgrading,wang2024conditional,li2023comparison}. To overcome the limitations of the initial generation, we introduced the Lane Diffusion module, which refines lane information through multi-step noise addition and denoising processes. Noise is added to simulate uncertainties, and denoising optimizes lane information by learning traffic state constraints, improving inference accuracy and model generalization. Constraints ensure the generated lane information aligns with real-world traffic patterns.

\subsection{Road-Lane Autoencoder-Decoder}
This module consists of a road-level encoder and a lane-level decoder to model spatio-temporal dependencies. The encoder constructs a static adjacency matrix from road topology and captures dynamic relationships using attention mechanisms, extracting high-dimensional representations. The decoder uses lane network topology and road-lane correspondences to reconstruct initial lane information, establishing spatio-temporal dependencies. The encoder captures complex road network relationships, and the decoder translates these into detailed lane information.

\subsubsection{Road-Level Encoder}

Relying solely on static adjacency relationships is insufficient to capture spatial dependencies \cite{li2017diffusion,zheng2020gman,wu2019graph}. Thus, we combine the attention mechanism with the GCN \cite{bruna2014spectral} to dynamically identify spatially similar road segments, even if not directly adjacent. This method more accurately captures complex spatial relationships, ensuring comprehensive intermediate vectors.

First, we use static convolution to process the static topology of the road network. Through the static convolution operation, we initially extract the structural information of the road network:

\begin{equation}
    x^{r_i,(m+1)} = \sigma \left( D^{-\frac{1}{2}}_R A^R D^{-\frac{1}{2}}_R x^{r_i,(m)} W^{(m)} \right) ,
\end{equation}
where \( x^{r_i,(m)} \) represents the feature matrix of road segment  $r_i$ at layer \( m \), \( D_R \) is the degree matrix, \( W^{(m)} \) is the weight matrix at layer \( m \), and \( \sigma \) is the activation function.

However, static convolution can only capture the fixed relationships between road segments and cannot dynamically adapt to changes in actual traffic conditions. To address this issue, we introduce an attention mechanism to dynamically adjust adjacency relationships. Through the attention mechanism, we can dynamically evaluate and adjust the relationship weights between each pair of nodes:
\begin{equation}
    \alpha_{r_i,r_j} = \frac{\exp(\text{LeakyReLU} \left( a^T \left[ W x^{r_i} \| W x^{r_j} \right] \right) )}{\sum_{r_k \in \mathcal{N}(r_i)} \exp(\text{LeakyReLU} \left( a^T \left[ W x^{r_i} \| W x^{r_k} \right] \right) )} ,
\end{equation}
where \( a \) is a learnable parameter vector, \( W \) is the weight matrix, \( \| \) denotes the concatenation operation, and \( \alpha_{r_i,r_j} \) is the normalized attention coefficient between road segment $r_i$ and road segment $r_j$.

By combining the information from static neighbor nodes and dynamically adjacent nodes, we can obtain comprehensive spatial feature vectors:
\begin{equation}
    \overline{x}^{r_i} = \beta \left( D^{-\frac{1}{2}} A^R D^{-\frac{1}{2}} x^{r_i} W \right) + (1 - \beta) \left( \sum_{r_j \in \mathcal{N}(r_i)} \alpha_{r_i,r_j} W x^{r_j} \right) ,
\end{equation}
where \( \beta \) is a balancing parameter used to adjust the weights of static and dynamic features. Through this combination, we can not only capture the fixed topological relationships within the road network but also flexibly adapt to the dynamic changes in actual traffic conditions.

After obtaining the comprehensive spatial feature vectors, we further consider spatio-temporal modeling factors. Existing studies have used multilayer perceptrons (MLPs) and simple channel-independent or channel-related strategies to segment sequences into patches \cite{wang2023st,vijay2023tsmixer,zeng2023transformers,gong2023patchmixer}, achieving superior performance in long-term predictions compared to complex models like Transformers\cite{vaswani2017attention,zhou2021informer}. This inspired us to design a simple structure for spatio-temporal modeling. However, unlike these methods, we do not use patch segmentation and channel strategies but instead use MLP combined with the similarity of node information in the comprehensive spatial vectors for spatio-temporal modeling to obtain intermediate vectors.

Specifically, we designed a temporal attention mechanism to dynamically adjust the spatio-temporal features by calculating the similarity and dynamic weights between nodes at different time segments:
\begin{equation}
    e^{r_i,r_j}_{(t)} = \text{ReLU} \left( \frac{W_q \overline{x}^{r_i}_t \cdot W_k \overline{x}^{r_j}_t}{\sqrt{d_k}} \right) 
\end{equation}

\begin{equation}
    \alpha^{r_i,r_j}_{(t)} = \frac{\exp(e^{r_i,r_j}_{(t)})}{\sum_{r_k \in \mathcal{N}(r_i)} \exp(e^{r_i,r_k}_{(t)})},
\end{equation}
where \(  e^{r_i,r_j}_{(t)} \) represents the attention coefficient between node \( r_i \) and node \( r_j \) between timestamps \( t \) and \( t-1 \), \(  \alpha^{r_i,r_j}_{(t)} \) is the normalized attention coefficient at timestamps \( t \), and \( d_k \) is the dimension of the key for scaling.

Through this temporal attention mechanism, we can effectively capture the temporal dependencies between nodes and incorporate them into the spatio-temporal features. The spatio-temporal feature update formula combined with the attention weights is as follows:

\begin{equation}
    \widetilde{x}_t^{r_i} = \sum_{r_j \in \mathcal{N}(r_i)} \alpha^{r_i,r_j}_{(t)} W_v \overline{x}^{r_j}_{t-1},
\end{equation}
where \( W_v \) is the weight matrix for the values, and \(  \widetilde{x}_t^{r_i} \) is the spatio-temporal feature vector at timestamp \( t \) updated through the attention mechanism.

Next, we process the updated spatio-temporal feature vectors for all time segments \( t \) through an MLP, further integrating node information to obtain the final intermediate hidden vector:
\begin{equation}
    \widehat{X}^R = \text{MLP}\left( \sum_{t=1}^T \widetilde{x}_t^{r_i} \right), 
\end{equation}
where \( \widehat{X}^R \) is the intermediate hidden vector processed by the MLP.

\subsubsection{Lane-Level Decoder}

The Lane-Level Decoder leverages the topology of the lane network and the correspondence between roads and lanes to decompose the intermediate hidden vector \( \widehat{X}^R \) obtained from the Road Encoder into corresponding lane information. First, to map \( \widehat{X}^R \) to the initial feature representation \( \ddot{x}^{l} \) of each lane segment in the lane network, we need to consider the correspondence between roads and lanes. Since each road segment corresponds to multiple lane segments, we map \( \widehat{X}^R\) to the initial feature representation of the lane segments using a weighted sum approach:
\begin{equation}
    \ddot{x}^{l_{i,j}} = W_d \widehat{x}^{r_i} + b_d ,
\end{equation}
where \( W_d \) is the weight matrix, and \( b_d \) is the bias term. Next, we use a structure similar to the encoder to enrich the similarity and differences of lane segments by utilizing the features of spatially and spatio-temporally adjacent nodes. This process can be viewed as an aggregation of relevant information. First, by combining the information from static and dynamic neighboring nodes, we obtain enriched lane-level spatial feature vectors:
\begin{small} 
\begin{equation}
    \overline{x}^{l_{i,j}} = \gamma \left( D_L^{-\frac{1}{2}} A^L D_L^{-\frac{1}{2}} \ddot{x}^{l_{i,j}} W_L \right) + (1 - \gamma) \left( \sum_{l_{a,b} \in \mathcal{N}(l_{i,j})} \alpha_{l_{i,j},l_{a,b}} W_L \ddot{x}^{l_{a,b}} \right) ,
\end{equation}
\end{small}
where \( \gamma \) is a balancing parameter. Subsequently, in the spatio-temporal dependency modeling, we consider the diffusion effects across different time periods. By aggregating information from relevant lane segments across different time dimensions with different weights, we can better restore the spatio-temporal dependencies between lanes:
\begin{equation}
    \widetilde{x}_t^{l_{i,j}} = \sum_{l_{a,b} \in \mathcal{N}(l_{i,j})} \alpha_{l_{i,j},l_{a,b}}^{(t)} W_{Lv} \overline{x}^{l_{i,j}}_{t-1},
\end{equation}
where \( W_{Lv} \) is the weight matrix for the values. Finally, we process the updated spatio-temporal feature vectors for all time segments \( t \) through an MLP, further integrating node information to obtain the final lane information:
\begin{equation}
    \widehat{X}^L = \text{MLP}\left( \sum_{t=1}^T \widetilde{x}_t^{l_{i,j}} \right),
\end{equation}
where \( \widehat{X}^L \) represents the initial lane information.

\subsection{Lane Diffusion Module}

While the Road-Lane Correlation Autoencoder-Decoder module provides an initial inference of lane information, Variational Autoencoders (VAEs) face challenges such as overfitting, posterior collapse, and poor sample quality \cite{xiao2023upgrading,wang2024conditional,li2023comparison}. Diffusion-based generation addresses these issues by providing robust approximations, avoiding posterior collapse, and producing higher-quality outputs \cite{ho2020denoising}. Following the Autoencoder-Decoder module, we introduce noise to simulate uncertainty and iteratively optimize lane information using traffic state constraints for more accurate inference.

\subsubsection{Forward Diffusion Process}

The forward diffusion process gradually adds noise to the initial lane information \( \widehat{X}^L \), incorporating road-lane interactions for better physical interpretability. This process is represented as a Markov chain:
\begin{equation}
     q(\widehat{X}^L_{(n)} | \widehat{X}^L_{(n-1)}, \widehat{X}^R_{(n-1)}) = 
     \mathcal{N}\Big(\widehat{X}^L_{(n)}; \sqrt{1-\beta_n} \, \widehat{X}^L_{(n-1)} + \gamma_n \, \widehat{X}^R_{(n-1)}, \, \beta_n \mathbf{I} \Big),
\end{equation}
where \(\widehat{X}^R_{(n-1)}\) denotes road information at step \( n-1 \), and \(\gamma_n\) is a weight controlling road influence. The lane information is progressively overlaid with random noise and road dynamics, simulating mutual dependencies in traffic.The noise addition follows:
\begin{equation}
    \widehat{X}^L_{(n)} = \sqrt{1 - \beta_n} \widehat{X}^L_{(n-1)} + \gamma_n \widehat{X}^R_{(n-1)} + \sqrt{\beta_n} \epsilon, \quad \epsilon \sim \mathcal{N}(0, \mathbf{I})
\end{equation}
where \(\epsilon\) represents random traffic disturbances, while \(\gamma_n \widehat{X}^R_{(n-1)}\) models road influence. This approach combines noise accumulation with road-lane dynamics, aligning the generation process with real-world traffic patterns.

\subsubsection{Reverse Diffusion Process}

The reverse diffusion process removes noise and reconstructs lane information using a probabilistic model \( p_\theta(\widehat{X}^L_{(n-1)} | \widehat{X}^L_{(n)}, \widehat{X}^R_{(n)}) \). This step refines the lane-road dependencies while denoising:
\begin{equation}
   p_\theta(\widehat{X}^L_{(n-1)} | \widehat{X}^L_{(n)}, \widehat{X}^R_{(n)}) 
    =  \, \mathcal{N}\Big(\widehat{X}^L_{(n-1)}; \, \mu_\theta(\widehat{X}^L_{(n)}, \widehat{X}^R_{(n)}, n), \Sigma_\theta(\widehat{X}^L_{(n)}, n) \Big)
\end{equation}

To capture the dependencies between lane and road information, we devised the reverse diffusion formula:
\begin{equation}
\begin{split}
    \widehat{X}^L_{(n-1)} = & \, \frac{1}{\sqrt{1 - \beta_n}} \left( \widehat{X}^L_{(n)} - \frac{\beta_n}{\sqrt{1 - \beta_n}} \epsilon_\theta(\widehat{X}^L_{(n)}, \widehat{X}^R_{(n)}, n) \right) \\
    & + \gamma_n \widehat{X}^R_{(n)} + \Sigma_\theta(\widehat{X}^L_{(n)}, n),
\end{split}
\end{equation}
where \(\epsilon_\theta(\widehat{X}^L_{(n)}, \widehat{X}^R_{(n)}, n)\) predicts noise, and \(\gamma_n \widehat{X}^R_{(n)}\) adjusts lane data based on road dynamics. This ensures that reconstructed lane information aligns with real traffic interactions.

\subsubsection{Incorporating Constraints}

To enhance the Lane Diffusion module's reliability, traffic state constraints, including speed and flow consistency, are applied during reverse diffusion to ensure adherence to real-world traffic rules. Lane information is refined iteratively using:
\begin{equation}
    \widehat{X}^L_{(n)} = \hat{X}^L_{(n)} - \eta \frac{\partial \mathcal{L}_{con}}{\partial \hat{X}^L_{(n)}}, 
\end{equation}
where \( \widehat{X}^L_{(n)} \) is the refined lane information, \( \eta \) is the learning rate, and \( \mathcal{L}_{con} \) is the constraint loss:
\begin{equation}
     \mathcal{L}_{con} = 
\begin{cases} 
\sum_{i=1}^I \left( x^{r_i} - \frac{1}{J_i} \sum_{j=1}^{J_i} x^{l_{i,j}} \right)^2 & \text{If } x^{r_i},x^{l_{i,j}} \text{are speed.}\\
 \sum_{i=1}^I \left( x^{r_i} - \sum_{j=1}^{J_i} x^{l_{i,j}} \right)^2 & \text{If } x^{r_i}, x^{l_{i,j}} \text{are flow.}
\end{cases}
\label{eq: constraint}
\end{equation}

These constraints ensure that generated lane information adheres to fundamental traffic principles. By combining constraints with diffusion, the Lane Diffusion module effectively captures traffic dynamics while maintaining strict compliance with physical laws, enhancing robustness and accuracy.

\subsection{Training Strategy}

The objective of training the RoadDiff framework is to ensure that the generated lane information aligns as closely as possible with the real lane information while satisfying various constraints. To achieve this, the loss function integrates reconstruction error and constraint error, including the following components:

\noindent\textbf{ Reconstruction Error:} This part of the loss ensures that the generated lane information matches the real lane information as closely as possible. The specific calculation formula is as follows:
\begin{equation}
    \mathcal{L}_{recon} = \mathbb{E}_{X^L, \epsilon, n} [\|\epsilon - \epsilon_\theta(\widehat{X}^L_n, n)\|^2], 
\end{equation}
where \( \epsilon \) represents the noise added to the data, and \( \epsilon_\theta \) represents the noise predicted by the model.

\noindent\textbf{Constraint Error:} $ \mathcal{L}_{con}$ ensures that the generated lane information meets the speed and flow constraints. 

\noindent\textbf{KL Divergence:} This part of the loss ensures that the reverse diffusion process approximates the inverse of the forward diffusion process as closely as possible. The specific formula is as follows:
\begin{equation}
    \mathcal{L}_{KL} = \mathbb{E}_q [\sum_{n=1}^N D_{KL}(q(\widehat{X}^L_{n-1} | \widehat{X}^L_n, \widehat{X}^L_0) \| p_\theta(\widehat{X}^L_{n-1} | \widehat{X}^L_n))], 
\end{equation}
where \( D_{KL} \) represents the KL divergence, which measures the difference between two distributions.

The combined loss function introduces uncertainty while progressively denoising and optimizing the lane information, ensuring that the generated lane information is both accurate and compliant with real-world traffic constraints. The combined loss function is formulated as follows:
\begin{equation}
    \mathcal{L}(\theta) = \mathcal{L}_{KL} + \mathcal{L}_{recon} + \lambda \mathcal{L}_{con},
\end{equation}
where \(\lambda\) is a balancing parameter used to adjust the weights of the different loss components, and $\theta$ represents the ensemble of adjustable parameters within the Roaddiff framework. To provide a clear and concise understanding of the RoadDiff model's training process, we outline its step-by-step strategy in \textbf{Algorithm 1}.

\begin{algorithm}[t]
\caption{Training Strategy for RoadDiff Model}
\KwIn{Road network topology $G^R$, lane network topology $G^L$, road network data $X^R$, time steps $T$, learning rate $\eta$, diffusion steps $\mathbb{N}$}
\KwOut{Trained RoadDiff model parameters $\theta$}

\BlankLine
\textbf{Initialization:} Initialize model parameters $\theta$.

\BlankLine
\textbf{Step 1: Road Network Encoding} \\
\For{each road segment $r_i \in G^R$}{
    \textbf{Graph Convolution:} \\
    $H^{(l+1)} = \sigma \left( D^{-\frac{1}{2}} A D^{-\frac{1}{2}} H^{(l)} W^{(l)} \right)$ \\
    \textbf{Attention Mechanism:} \\
    $e_{ij} = \text{LeakyReLU}\left( a^T \left[ W H_i \| W H_j \right] \right)$ \\
    $\alpha_{ij} = \frac{\exp(e_{ij})}{\sum_{k \in \mathcal{N}(i)} \exp(e_{ik})}$ \\
    Aggregate enriched features: \\
    $H' = \beta \left( D^{-\frac{1}{2}} A D^{-\frac{1}{2}} H W \right) + (1 - \beta) \left( \sum_{j \in \mathcal{N}(i)} \alpha_{ij} W H_j \right)$
}

\BlankLine
\textbf{Step 2: Lane Network Decoding} \\
\For{each lane segment $l_{i,j} \in G^L$}{
    \textbf{Mapping Road to Lane:} \\
    $H^{(0)}_{i,j} = W_d Z_i + b_d$ \\
    \textbf{Graph Convolution and Attention for Lanes:} Similar to Step 1, considering lane-specific topology.
}

\BlankLine
\textbf{Step 3: Lane Diffusion Module} \\
\For{$n = 1, 2, \dots, \mathbb{N}$}{
    \textbf{Forward Diffusion:} \\
    $\widehat{X}^L_{(n)} = \sqrt{1 - \beta_n} \widehat{X}^L_{(n-1)} + \gamma_n \widehat{X}^R_{(n-1)} + \sqrt{\beta_n} \epsilon, \quad \epsilon \sim \mathcal{N}(0, \mathbf{I})$ \\
    \textbf{Reverse Diffusion:} \\
    $\widehat{X}^L_{(n-1)} = \frac{1}{\sqrt{1 - \beta_n}} \left( \widetilde{X}^L_{(n)} - \frac{\beta_n}{\sqrt{1 - \beta_n}} \epsilon_\theta(\widetilde{X}^L_{(n)}, n) \right) + \gamma_n \widehat{X}^R_{(n)} + \Sigma_\theta(\widetilde{X}^L_{(n)}, n)$
}

\BlankLine
\textbf{Step 4: Loss Computation and Optimization} \\
Compute total loss $\mathcal{L}(\theta)$:
\[
\mathcal{L}(\theta) = \mathcal{L}_{KL} + \mathcal{L}_{recon} + \lambda \mathcal{L}_{con}
\]
Update model parameters $\theta$ using Adam optimizer with learning rate $\eta$.

\BlankLine
\textbf{Step 5: Early Stopping} \\
Apply early stopping based on validation performance to prevent overfitting.

\BlankLine
\Return{Trained RoadDiff model parameters $\theta$}
\end{algorithm}

\section{Experiment}
The primary objective of the RoadDiff framework is to provide a novel solution for spatio-temporal graph node generation problems and fine-grained road traffic inference problems. We evaluated RoadDiff's capabilities using six key datasets of road-lane speed and flow from two different sources. These datasets encompass various road types, including freeways, urban expressways, and roads with irregular lane counts. 

% \vspace{-0.5\baselineskip}
\subsection{Datasets}
We utilized datasets from two main sources: PeMS and the South China (HuaNan) Expressway in China, representing freeways and urban expressways, respectively.
The PeMS dataset originates from the Santa Ana Freeway in Los Angeles, USA, and includes data from eight sensors. This dataset is part of the well-known Performance Measurement System(PeMS) public dataset, comprising traffic speed and flow information from February 5 to March 5, 2017, covering roads with five or six lanes. The dataset with only five-lane roads is referred to as the PeMS dataset, while the dataset with irregular lane counts is referred to as PeMS\_F, with lane counts distributed as [5, 6, 5, 6, 5, 6, 5, 5]. 
% These two datasets offer practical scenarios for assessing fine-grained traffic inference methods under both uniform and heterogeneous lane configurations. 

Additionally, we used traffic flow and speed data from 18 sensors on the South China Expressway in Guangzhou, China. This dataset was collected over 30 days from July 22 to August 22, 2022. The statistical information of the datasets is provided in \textbf{Table~\ref{tab:datasets}}.

\begin{table}[h]
  \centering
  \vspace{-0.25cm}
  \caption{Statistic of Datasets}
  \vspace{-0.25cm}
  \resizebox{0.9\linewidth}{!}{

    \begin{tabular}{cccc}
    \toprule
    Parameters  & PeMS & PeMS\_F & HuaNan \\
    \midrule
    Timespan & 2/5/2017-3/5/2017                                                     & 2/5/2017-3/5/2017                                                     & 7/22/2022-8/22/2022 \\
    Region & Los Angeles & Los Angeles & Guangzhou \\
    \# Objects & 40(8) & 43(8) & 72(18) \\
    Road Type & Freeway & Freeway & Urban Expressway \\
    \# Timestamps & 8059 & 8059 & 44640 \\
    Unit(Speed) & Miles/Hour & Miles/Hour & Kilometers /Hour \\
    Unit(Flow) & Veh/5-Min & Veh/5-Min & Veh/2-Min \\
    The Time Interval & 5 min & 5 min & 2min \\
    \bottomrule
    \end{tabular}%
    }
  \label{tab:datasets}%
  \vspace{-0.35cm}
\end{table}%

\subsection{Evaluation Metrics}
To ensure a fair evaluation of the prediction performance, we employed three commonly used metrics in traffic prediction: Mean Absolute Error (MAE), Root Mean Squared Error (RMSE), and Mean Absolute Percentage Error (MAPE). The formulas for these metrics are detailed in \textbf{Appendix~\ref{app:metrics}}. The first two metrics measure absolute prediction errors, while the last metric measures relative prediction errors. For all these metrics, smaller values indicate better predictive performance. 
\begin{table*}[t]
  \centering
  \vspace{-0.15cm}
  \caption{Comparison on PeMS Dataset}
  \vspace{-0.25cm}
    \resizebox{0.9\linewidth}{!}{
    \begin{tabular}{ccccccccccc|ccccccccc}
    \toprule
    \multicolumn{1}{c}{\multirow{2}[4]{*}{Model}} & \multicolumn{2}{c}{Traffic Type} &     &     & \multicolumn{3}{c}{ Traffic Speed} &     &     & \multicolumn{1}{r}{} & \multicolumn{9}{c}{ Traffic Flow} \\
\cmidrule{2-20}        & \multicolumn{2}{c}{Time Windows} & 1   &     & \multicolumn{3}{c}{3} & \multicolumn{3}{c|}{6} & \multicolumn{3}{c}{1} & \multicolumn{3}{c}{3} & \multicolumn{3}{c}{6} \\
\cmidrule{2-20}    Types & \multicolumn{1}{l}{Metrics} & MAE & RMSE & MAPE & MAE & RMSE & MAPE & MAE & RMSE & MAPE & MAE & RMSE & MAPE & MAE & RMSE & MAPE & MAE & RMSE & MAPE \\
    \midrule
    \textbf{Basic} & Physics & $\underline{11.47}$ & $\underline{14.78}$ & $\underline{50.78\%}$ & $\underline{11.27}$ & $\underline{14.87}$ & $\underline{50.69\%}$ & $\underline{11.37}$ & $\underline{14.77}$ & $\underline{50.57\%}$ & $\underline{19.35}$ & $\underline{32.69}$ & $\underline{47.87\%}$ & $\underline{19.41}$ & $\underline{32.81}$ & $\underline{50.67\%}$ & $\underline{20.28}$ & $\underline{32.05}$ & $\underline{36.19\%}$ \\
    \midrule
    \multicolumn{1}{c}{\multirow{5}[2]{*}{\begin{sideways}\textbf{FUFI}\end{sideways}}} & UrbanFM & 12.21  & 15.24  & 60.46\% & 12.06  & 15.15  & 59.95\% & 10.77  & 17.17  & 58.21\% & 16.83  & 28.08  & 26.13\% & 16.78  & 27.99  & 26.11\% & 16.70  & 27.90  & 26.05\% \\
        & FODE & 12.03  & 15.48  & 71.24\% & 11.88  & 15.39  & 70.64\% & 10.61  & 17.43  & 68.59\% & 16.58  & 28.51  & 30.79\% & 16.54  & 28.42  & 30.76\% & 16.46  & 28.34  & 30.70\% \\
        & UrbanPy & 11.99  & 14.95  & 59.37\% & 11.84  & 14.86  & 58.88\% & 10.58  & 16.83  & 57.16\% & 16.53  & 27.53  & 25.66\% & 16.48  & 27.45  & 25.64\% & 16.41  & 27.36  & 25.58\% \\
        & DeepLGR & 12.77  & 15.52  & 61.18\% & 12.61  & 15.42  & 60.67\% & 11.26  & 15.47  & 58.90\% & 17.60  & 28.58  & 26.44\% & 17.55  & 28.49  & 26.42\% & 17.47  & 28.40  & 26.36\% \\
        & CUFAR & $\underline{10.95}$ & $\underline{11.43}$ & $\underline{54.09\%}$ & $\underline{10.81}$ & $\underline{11.36}$ & $\underline{53.64\%}$ & $\underline{10.66}$ & $\underline{12.87}$ & $\underline{52.08\%}$ & $\underline{15.09}$ & $\underline{25.98}$ & $\underline{23.38\%}$ & $\underline{15.05}$ & $\underline{25.90}$ & $\underline{23.36\%}$ & $\underline{14.98}$ & $\underline{25.82}$ & $\underline{23.31\%}$ \\
    \midrule
    \multicolumn{1}{c}{\multirow{4}[2]{*}{\begin{sideways}\textbf{ED}\end{sideways}}} & DCRNN & 14.89  & 17.16  & 64.67\% & 14.80  & 17.09  & 64.46\% & 14.80  & 17.08  & 64.25\% & 22.05  & 37.01  & 41.67\% & 22.39  & 37.35  & 42.18\% & 22.38  & 37.34  & 42.13\% \\
        & GMAN & 13.79  & 15.31  & 59.81\% & 13.71  & 15.24  & 59.61\% & 13.71  & 15.24  & 59.43\% & 20.42  & 33.01  & 38.53\% & 20.74  & 33.31  & 39.01\% & 20.72  & 33.30  & 38.96\% \\
        & Bi-STAT & 12.79  & 14.91  & 53.98\% & 12.72  & 14.85  & 53.80\% & 12.72  & 14.84  & 53.63\% & 18.95  & 32.16  & 34.78\% & 19.24  & 32.46  & 35.21\% & 19.23  & 32.45  & 35.16\% \\
        & MegaCRN & $\underline{11.25}$ & $\underline{13.82}$ & $\underline{45.19\%}$ & $\underline{11.18}$ & $\underline{13.76}$ & $\underline{45.04\%}$ & $\underline{11.18}$ & $\underline{13.75}$ & $\underline{44.90\%}$ & $\underline{16.66}$ & $\underline{29.80}$ & $\underline{29.12\%}$ & $\underline{16.92}$ & $\underline{30.07}$ & $\underline{29.48\%}$ & $\underline{16.91}$ & $\underline{30.06}$ & $\underline{29.44\%}$ \\
    \midrule
    \multicolumn{1}{c}{\multirow{7}[2]{*}{\begin{sideways}\textbf{STG + Linear}\end{sideways}}} & STGCN & 14.85  & 20.99  & 63.39\% & 14.77  & 20.90  & 63.18\% & 14.77  & 20.89  & 62.98\% & 22.00  & 40.74  & 40.84\% & 22.34  & 41.11  & 41.35\% & 22.33  & 41.09  & 41.29\% \\
        & MTGNN & $\underline{10.87}$ & $\underline{13.94}$ & $\underline{39.12\%}$ & $\underline{10.81}$ & $\underline{13.88}$ & $\underline{38.99\%}$ & $\underline{10.81}$ & $\underline{13.87}$ & $\underline{38.87\%}$ & 16.48  & 27.50  & 25.59\% & 16.44  & $\underline{27.42}$ & $\underline{25.57\%}$ & $\underline{16.36}$ & $\underline{27.33}$ & $\underline{25.52\%}$ \\
        & ASTGCN & 14.34  & 17.72  & 51.53\% & 14.26  & 17.65  & 51.35\% & 14.26  & 17.64  & 51.19\% & 21.75  & 34.98  & 33.71\% & 21.70  & 34.87  & 33.68\% & 21.59  & 34.76  & 33.61\% \\
        & GraphWaveNet & 11.54  & 14.86  & 40.84\% & 11.47  & 14.80  & 40.71\% & 11.47  & 14.80  & 40.58\% & 17.49  & 29.33  & 26.72\% & 17.45  & 29.25  & 26.70\% & 17.36  & 29.15  & 26.64\% \\
        & STSGCN & 11.18  & 14.12  & 39.55\% & 11.11  & 14.06  & 39.41\% & 11.11  & 14.06  & 39.29\% & 16.56  & $\underline{27.41}$ & 25.48\% & 16.81  & 27.66  & 25.79\% & 16.80  & 27.65  & 25.76\% \\
        & AGCRN & 11.13  & 14.17  & 39.72\% & 10.87  & 13.94  & 39.07\% & 10.82  & 13.89  & 38.92\% & $\underline{16.24}$ & 27.49  & $\underline{25.39\%}$ & $\underline{16.38}$ & 27.99  & 33.18\% & 16.37  & 27.40  & 25.57\% \\
        & STGODE & 12.70  & 17.60  & 47.28\% & 12.63  & 17.53  & 47.12\% & 12.63  & 17.52  & 46.98\% & 18.82  & 34.17  & 30.46\% & 19.11  & 34.48  & 30.84\% & 19.10  & 34.47  & 30.80\% \\
    \midrule
    \textbf{Ours} & RoadDiff & \textit{\textbf{7.04 }} & \textit{\textbf{9.22 }} & \textit{\textbf{21.62\%}} & \textit{\textbf{7.04 }} & \textit{\textbf{9.22 }} & \textit{\textbf{21.47\%}} & \textit{\textbf{7.05 }} & \textit{\textbf{9.23 }} & \textit{\textbf{21.48\%}} & \textit{\textbf{14.56 }} & \textit{\textbf{23.92 }} & \textit{\textbf{16.23\%}} & \textit{\textbf{14.51 }} & \textit{\textbf{23.82 }} & \textit{\textbf{16.20\%}} & \textit{\textbf{14.42 }} & \textit{\textbf{23.68 }} & \textit{\textbf{16.15\%}} \\
    \bottomrule
    \end{tabular}%
    }
    \vspace{-0.25cm}
    \label{tab:PeMS_main_results}%
\end{table*}%

\subsection{Baselines}

Given the lack of existing solutions for the spatio-temporal graph node generation problem, we adapted 17 baseline models from basic baseline (Physics), existing urban inference models, encoder-decoder framework models, and popular graph-based traffic prediction models to expand the potential solutions for this problem.

\noindent\textbf{Basic Baselines:} We designed a physics-based baseline without learning layers, mapping road speed and flow to lanes via replication and averaging, and calculating errors within a unified framework.

\noindent\textbf{Urban Inference Models (FUFI): }In the PeMS and HuaNan datasets, the lanes are regular. We processed them using grid-based methods, treating the input road-level data as an image with a width of 1 and a length equal to the number of road segments, with the time dimension as the channel dimension. The output is an image with a width equal to the number of lanes, a length equal to the number of road segments, and a time dimension as the channel. For the PeMS\_F dataset, which includes segments with different numbers of lanes, we padded the missing lanes with zeros to make it a regular rectangle. The urban inference models we used include UrbanFM \cite{liang2019urbanfm}, FODE \cite{zhou2020enhancing}, UrbanPy \cite{ouyang2020fine}, DeepLGR \cite{liang2021revisiting}, and CUFAR \cite{yu2023overcoming}.

\noindent\textbf{Encoder-Decoder Frameworks (ED) :} Encoder-decoder frameworks are suitable solutions for the spatio-temporal graph node generation problem. We only need to change the shape of the decoder output tensor to align it with the lane data to calculate the training loss. The encoder-decoder framework models we compared include DCRNN \cite{li2017diffusion}, GMAN \cite{zheng2020gman}, Bi-STAT \cite{chen2022bidirectional}, and MegaCRN \cite{jiang2023spatio}.

\noindent\textbf{Spatio-Temporal Graph Structures (STG) :} For existing graph-based models, we added a linear layer to their output to align the output shape with the lane-level data for calculating the training loss. The graph-based models we compared include STGCN \cite{yu2017spatio}, MTGNN \cite{yu2017spatio}, ASTGCN \cite{guo2019attention}, GraphWaveNet \cite{wu2019graph}, STSGCN \cite{song2020spatial}, AGCRN \cite{bai2020adaptive}, and STGODE \cite{fang2021spatial}.

All experiments were conducted on a computing platform equip-ped with Intel® Xeon® CPU Max 9462 processors @ 2.70 GHz and six NVIDIA A100 80Gb SXM GPUs. We used the Adam stochastic gradient descent algorithm for network training, with a learning rate of 1e-4, a batch size of 64, and a maximum of 1000 training iterations. To prevent overfitting, we employed an early stopping strategy based on validation set performance. Starting from the 20th training epoch, the learning rate was halved every 10 epochs. The main experimental setup had a diffusion step number of 10. All baseline models, except for the physics-based ones, were run using the authors' provided code and the Optuna library was employed to search for optimal hyperparameters, ensuring fair comparisons unaffected by task adaptation issues.

% \vspace{-0.35cm}
\subsection{Main Experimental Results}
\begin{table*}[htbp]
  \centering
  \vspace{-0.15cm}
  \caption{Comparison on PeMS\_F Dataset}
  \vspace{-0.25cm}
   \resizebox{0.9\linewidth}{!}{
     \begin{tabular}{lcccccccccc|ccccccccc}
    \toprule
    \multicolumn{1}{c}{\multirow{2}[4]{*}{Model}} & \multicolumn{2}{c}{Traffic Type} &     &     & \multicolumn{3}{c}{ Traffic Speed} &     &     & \multicolumn{1}{r}{} & \multicolumn{9}{c}{ Traffic Flow} \\
\cmidrule{2-20}        & \multicolumn{2}{c}{Time Windows} & 1   &     & \multicolumn{3}{c}{3} & \multicolumn{3}{c|}{6} & \multicolumn{3}{c}{1} & \multicolumn{3}{c}{3} & \multicolumn{3}{c}{6} \\
\cmidrule{2-20}    \multicolumn{1}{c}{Types} & \multicolumn{1}{l}{Metrics} & MAE & RMSE & MAPE & MAE & RMSE & MAPE & MAE & RMSE & MAPE & MAE & RMSE & MAPE & MAE & RMSE & MAPE & MAE & RMSE & MAPE \\
    \midrule
    \multicolumn{1}{c}{\textbf{Basic}} & Physics & $\underline{11.02}$ & $\underline{15.37}$ & $\underline{44.05\%}$ & $\underline{11.12}$ & $\underline{15.39}$ & $\underline{44.55\%}$ & $\underline{11.01}$ & $\underline{15.37}$ & $\underline{44.07\%}$ & $\underline{16.92}$ & $\underline{26.36}$ & $\underline{36.72\%}$ & $\underline{15.66}$ & $\underline{24.08}$ & $\underline{38.93\%}$ & $\underline{15.49}$ & $\underline{23.12}$ & $\underline{43.560\%}$ \\
    \midrule
    \multicolumn{1}{c}{\multirow{5}[2]{*}{\begin{sideways}\textbf{FUFI}\end{sideways}}} & UrbanFM & 14.74  & 18.50  & 82.46\% & 14.79  & 18.52  & 82.38\% & 14.74  & 18.35  & 76.02\% & 20.45  & 34.87  & 41.66\% & 20.43  & 34.60  & 36.93\% & 20.53  & 35.09  & 41.77\% \\
        & FODE & 15.30  & 19.78  & 92.09\% & 15.30  & 19.81  & 92.01\% & 15.31  & 19.73  & 91.96\% & 21.23  & 37.29  & 51.70\% & 21.21  & 37.01  & 45.83\% & 21.31  & 37.53  & 51.84\% \\
        & UrbanPy & 14.94  & 18.71  & 83.54\% & 14.54  & 18.73  & 83.46\% & 14.93  & 18.74  & 83.68\% & 20.72  & 35.27  & 42.21\% & 20.71  & 35.00  & 37.41\% & 20.80  & 35.49  & 42.32\% \\
        & DeepLGR & 15.91  & 19.42  & 86.08\% & 15.90  & 19.44  & 86.00\% & 15.86  & 19.32  & 81.15\% & 22.07  & 36.61  & 43.49\% & 22.05  & 36.33  & 38.55\% & 22.15  & 36.84  & 43.60\% \\
        & CUFAR & $\underline{13.50}$ & $\underline{17.48}$ & $\underline{75.32\%}$ & $\underline{13.49}$ & $\underline{17.50}$ & $\underline{75.25\%}$ & $\underline{12.95}$ & $\underline{17.19}$ & $\underline{72.12\%}$ & $\underline{18.73}$ & $\underline{32.94}$ & $\underline{38.05\%}$ & $\underline{18.71}$ & $\underline{32.69}$ & $\underline{33.73\%}$ & $\underline{18.80}$ & $\underline{33.15}$ & $\underline{38.15\%}$ \\
    \midrule
    \multicolumn{1}{c}{\multirow{4}[2]{*}{\begin{sideways}\textbf{ED}\end{sideways}}} & DCRNN & 13.91  & 17.55  & 55.12\% & 13.86  & 17.48  & 54.94\% & 13.88  & 17.52  & 54.85\% & 20.81  & 31.61  & 41.79\% & 20.84  & 31.62  & 41.77\% & 21.10  & 31.90  & 41.93\% \\
        & GMAN & 12.62  & 16.33  & 49.93\% & 12.57  & 16.27  & 49.77\% & 12.59  & 16.30  & 49.69\% & 18.87  & 27.62  & 37.86\% & 18.91  & 27.62  & 37.83\% & 19.14  & 27.87  & 37.98\% \\
        & Bi-STAT & 11.71  & 15.94  & 45.06\% & 11.66  & 15.88  & 44.92\% & 11.68  & 15.91  & 44.84\% & 17.51  & $\underline{26.91}$ & 34.17\% & 17.54  & 27.51  & 34.15\% & 17.75  & $\underline{27.15}$ & 34.28\% \\
        & MegaCRN & $\underline{10.94}$ & $\underline{15.28}$ & $\underline{44.77\%}$ & $\underline{10.93}$ & $\underline{15.25}$ & $\underline{44.70\%}$ & $\underline{10.88}$ & $\underline{15.20}$ & $\underline{44.52\%}$ & $\underline{16.36}$ & 27.52  & $\underline{33.94\%}$ & $\underline{16.44}$ & $\underline{27.49}$ & $\underline{33.98\%}$ & $\underline{16.54}$ & 27.69  & $\underline{34.04\%}$ \\
    \midrule
    \multicolumn{1}{c}{\multirow{7}[2]{*}{\begin{sideways}\textbf{STG + Linear}\end{sideways}}} & STGCN & 14.16  & 19.82  & 61.70\% & 14.15  & 19.79  & 61.60\% & 14.09  & 19.72  & 61.37\% & 21.18  & 35.70  & 46.78\% & 21.28  & 35.80  & 46.83\% & 21.42  & 35.92  & 46.91\% \\
        & MTGNN & 10.75  & 15.33  & 43.51\% & 10.70  & 15.26  & 43.37\% & 10.72  & 15.30  & 43.30\% & 16.35  & 27.67  & 33.06\% & 16.25  & 27.17  & 25.50\% & 16.29  & 27.84  & 33.15\% \\
        & ASTGCN & 13.64  & 17.73  & 55.26\% & 13.63  & 17.70  & 55.17\% & 13.57  & 17.65  & 54.96\% & 20.40  & 31.94  & 41.90\% & 20.51  & 32.03  & 41.95\% & 20.63  & 32.14  & 42.01\% \\
        & GraphWaveNet & $\underline{9.72}$ & $\underline{13.13}$ & $\underline{32.61\%}$ & $\underline{9.68}$ & $\underline{13.08}$ & $\underline{32.50\%}$ & $\underline{9.69}$ & $\underline{13.11}$ & $\underline{32.45\%}$ & $\underline{15.78}$ & $\underline{26.29}$ & $\underline{26.44\%}$ & $\underline{15.68}$ & $\underline{26.84}$ & $\underline{25.40\%}$ & $\underline{15.72}$ & $\underline{25.45}$ & $\underline{26.51\%}$ \\
        & STSGCN & 12.32  & 16.48  & 50.55\% & 12.31  & 16.45  & 50.47\% & 12.25  & 16.40  & 50.27\% & 18.42  & 29.68  & 38.32\% & 18.51  & 29.76  & 38.37\% & 18.62  & 29.86  & 38.43\% \\
        & AGCRN & 10.77  & 15.36  & 43.60\% & 10.76  & 15.34  & 43.53\% & 10.72  & 15.29  & 43.36\% & 16.11  & 27.67  & 33.06\% & 16.19  & 27.74  & 33.09\% & 16.29  & 27.84  & 33.15\% \\
        & STGODE & 10.84  & 14.37  & 38.44\% & 10.80  & 14.31  & 38.32\% & 10.82  & 14.35  & 38.25\% & 16.21  & 26.89  & 29.15\% & 16.24  & 26.89  & 29.13\% & 16.44  & 26.13  & 29.24\% \\
    \midrule
    \textbf{Ours} & RoadDiff & \textit{\textbf{6.93 }} & \textit{\textbf{9.09 }} & \textit{\textbf{21.43\%}} & \textit{\textbf{6.95 }} & \textit{\textbf{9.11 }} & \textit{\textbf{21.54\%}} & \textit{\textbf{6.94 }} & \textit{\textbf{9.10 }} & \textit{\textbf{21.36\%}} & \textit{\textbf{14.99 }} & \textit{\textbf{25.51 }} & \textit{\textbf{20.67\%}} & \textit{\textbf{14.97 }} & \textit{\textbf{25.38 }} & \textit{\textbf{20.69\%}} & \textit{\textbf{14.83 }} & \textit{\textbf{24.56 }} & \textit{\textbf{16.28\%}} \\
    \bottomrule
    \end{tabular}%
    }
  \label{tab:PeMSF_main_results}%
  \vspace{-0.25cm}
\end{table*}%

In the main experiment, we used \textbf{bold} to highlight the overall best results and \underline{underlined} the best results within each category of baselines. Some experiments on the HuaNan dataset are presented in \textbf{Appendix~\ref{app:exp}}.

\textbf{Table \ref{tab:PeMS_main_results}} presents the performance comparison of the RoadDiff framework with various baseline models on the PeMS dataset, covering traffic speed and flow inference for 1, 3, and 6 time windows. The results indicate that, unlike prediction tasks, the length of the time window has minimal impact on inference errors, with only minor differences observed. In some cases, certain models even perform better with a 6-time window than with a 1-time window. For traffic speed, most baseline models performed worse than the physics-based baseline, indicating that while other baselines can be adapted to the FRTI problem, their effectiveness is limited. This highlights the strong physical constraints inherent to the FRTI problem. Compared to the best-performing baseline, RoadDiff reduced errors by approximately 40\% (calculated as the ratio of error reduction to RoadDiff’s error), demonstrating not only the effectiveness of RoadDiff in addressing the FRTI problem but also the problem's uniqueness and the necessity of designing specialized models for spatio-temporal graph node generation.
In the road-to-lane flow dataset, most baseline models outperformed the physics-based baseline, demonstrating that deep learning models can overcome the limitations of simple physical constraints when learning the significant variations in dependencies between road and lane flow. Among these, FUFI models performed well overall, with CUFAR emerging as the best-performing baseline, although its error was still higher than that of RoadDiff. Additionally, the physics-based baseline exhibited a larger MAPE, reflecting significant relative differences across lane data and increased errors when using average road flow as lane flow.

\textbf{Table \ref{tab:PeMSF_main_results}} further compares the models on the dataset with irregular lane counts. In the traffic speed dataset, RoadDiff continued to exhibit the best inference performance, while encoder-decoder (ED) and spatio-temporal graph-based models also showed improved accuracy. Compared to the regular datasets, FUFI baselines experienced a notable decline in performance on irregular lane data, primarily due to the inherent limitations of grid-based image models in directly modeling irregular lane structures. Meanwhile, spatio-temporal graph-based models, such as GraphWaveNet, MTGNN, and AGCRN, performed well on irregular lane data, highlighting their potential as effective solutions to the FRTI problem. In the traffic flow dataset, FUFI baselines similarly showed significant performance declines. Although RoadDiff's performance slightly decreased, it still maintained a marginal advantage over GraphWaveNet. Overall, spatio-temporal graph-based models stood out among all deep learning baselines, further validating the importance of effectively modeling spatio-temporal dependencies for accurate traffic flow inference.

% \vspace{-0.05cm}
\subsection{Study on Time Window Size} 
\begin{figure*} [t]
\vspace{-0.25cm}
\subfigure[PeMS Traffic Speed Dataset]{
\centering
\includegraphics[width=0.25\linewidth]{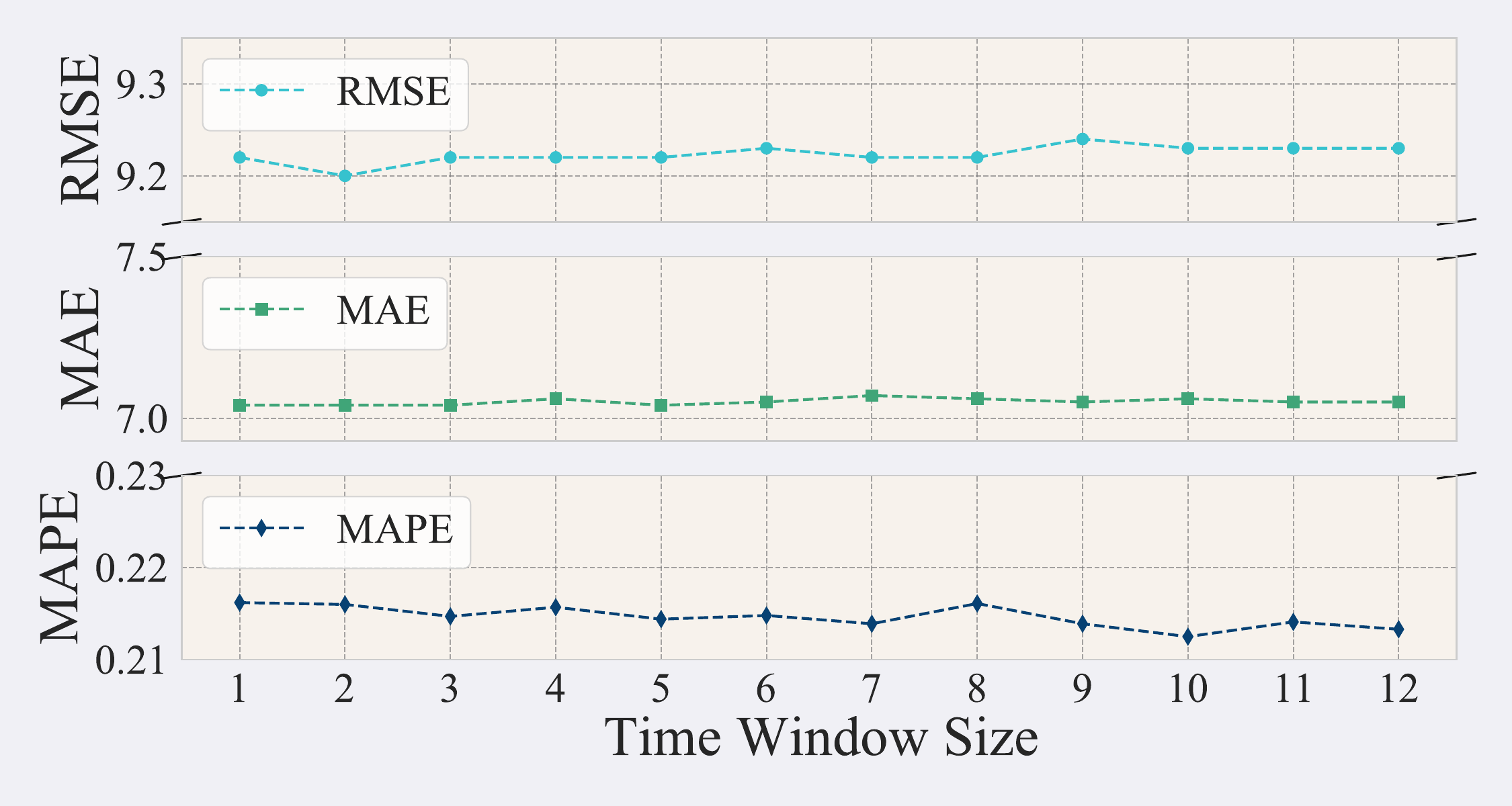}
}%
\subfigure[PeMS Traffic Flow Dataset]{
\centering
\includegraphics[width=0.25\linewidth]{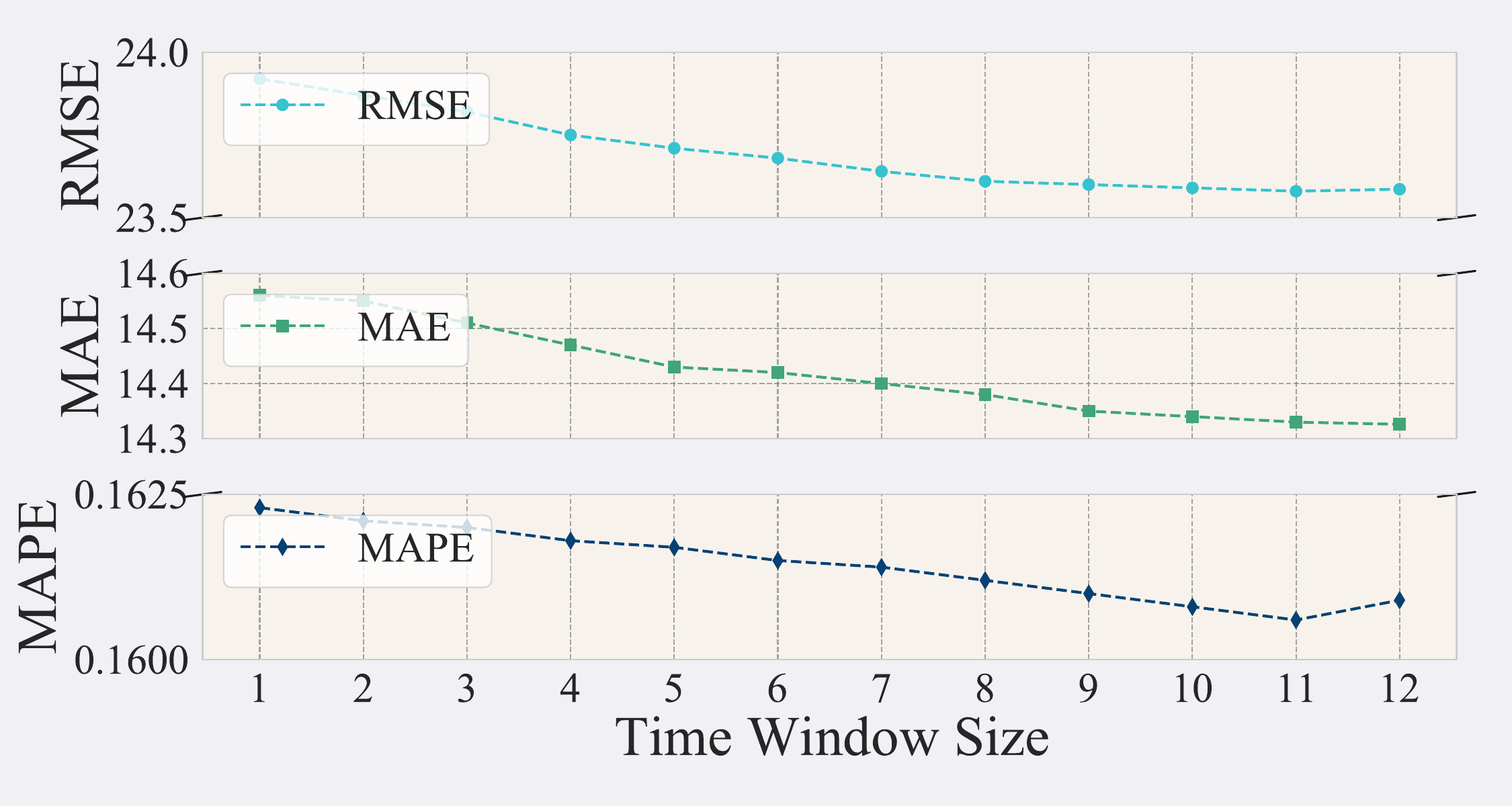}
}%
\subfigure[PeMS\_F Traffic Speed Dataset]{
\centering
\includegraphics[width=0.25\linewidth]{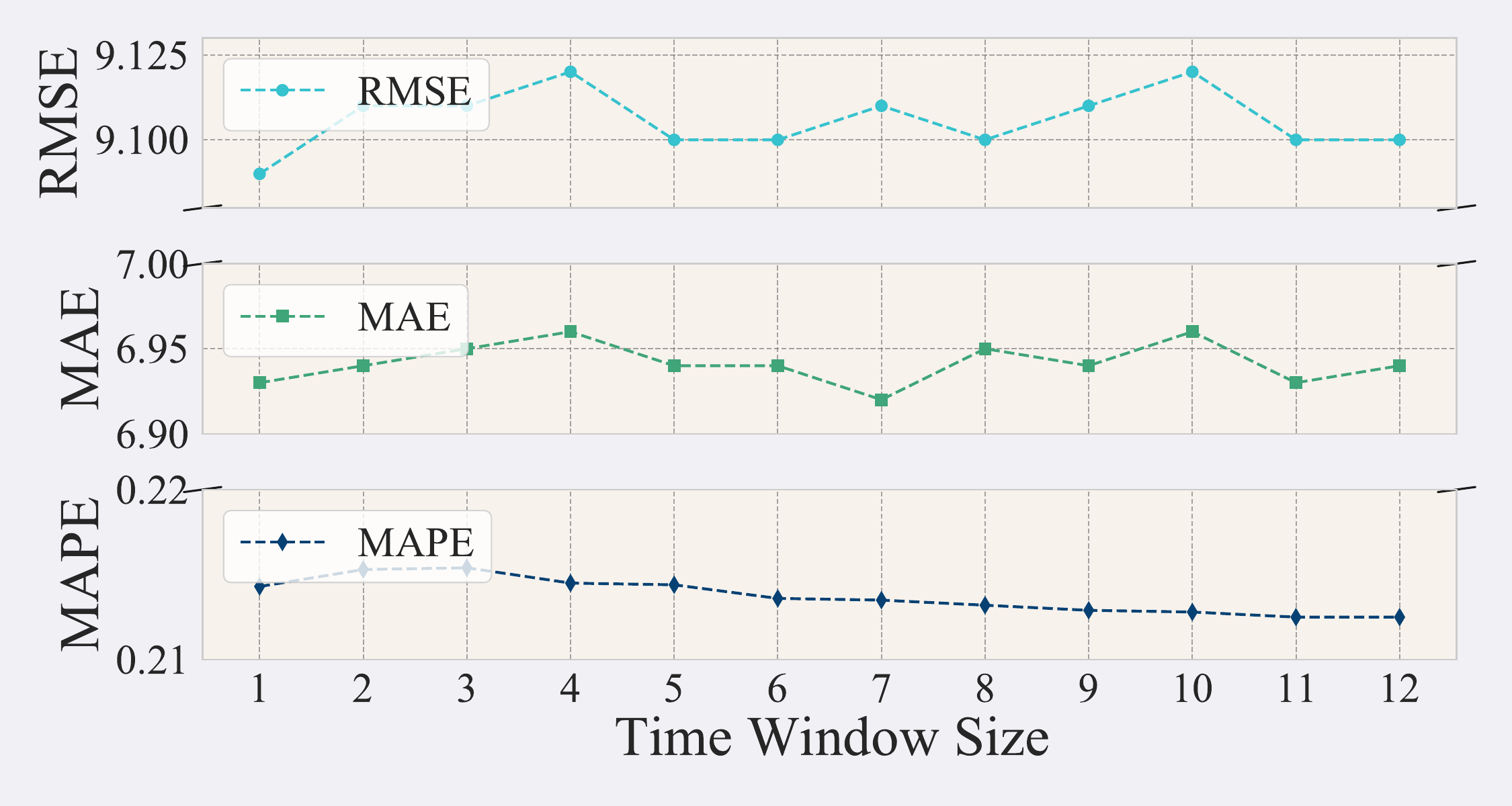}
}%
\subfigure[PeMS\_F Traffic Flow Dataset]{
\centering
\includegraphics[width=0.25\linewidth]{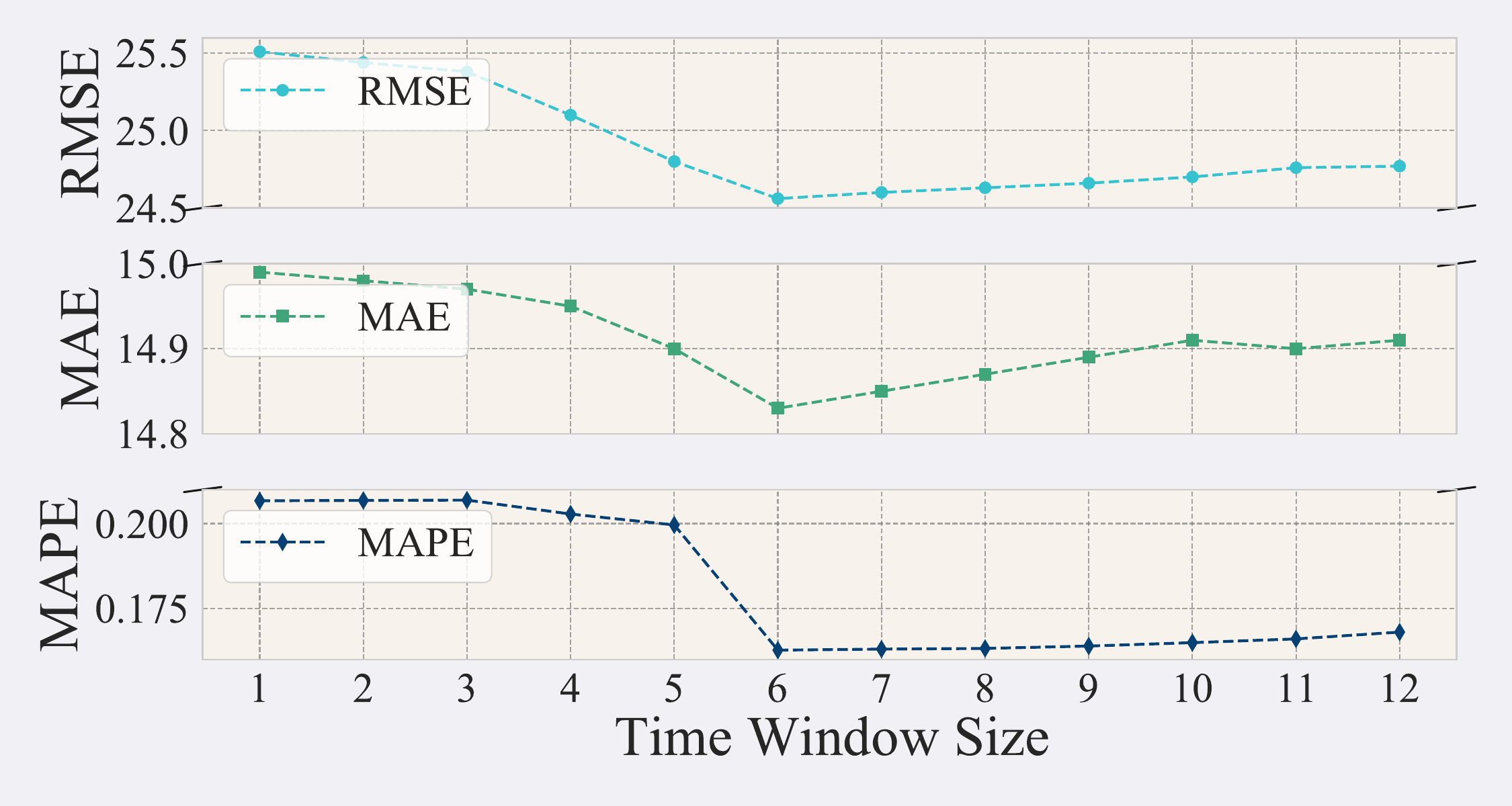}
}%
\centering
\vspace{-0.35cm}
\caption{The Variation in Inference Error under Different Time Window Sizes.}
\vspace{-0.25cm}
\label{fig: exp_time_window}
\end{figure*}

In \textbf{Figure \ref{fig: exp_time_window}}, we illustrate the changes in the three traffic speed and traffic flow metrics of RoadDiff on the PeMS and PeMS\_F datasets with longer time windows to study its ability to model spatio-temporal dependencies over extended periods. From subplots (a) and (c), it can be observed that the inference ability for traffic speed is minimally affected by fluctuations in the time window, and it even performs better over longer periods. This demonstrates RoadDiff's stability in spatio-temporal modeling of traffic speed. From subplots (b) and (d), it is evident that RoadDiff's performance improves with increasing time windows in the traffic flow dataset. Initially, RoadDiff performs poorly with smaller time windows, but as the time window lengthens, the inference error gradually decreases. This improvement is particularly notable in the PeMS\_F dataset, where the error significantly reduces over 4-6 time windows, highlighting the importance of capturing spatio-temporal relationships for inference, especially in traffic flow data where these dependencies are more pronounced.

\begin{figure} [h]
% \vspace{-0.40cm}
\includegraphics[width=0.99\linewidth]{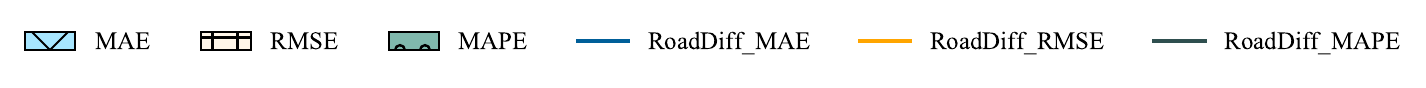} 
 % \vspace{-0.45cm} 
\newline
\centerline{
\subfigure[PeMS Traffic Speed]{
\vspace{-0.3cm}
\centering
\includegraphics[width=0.45\linewidth]{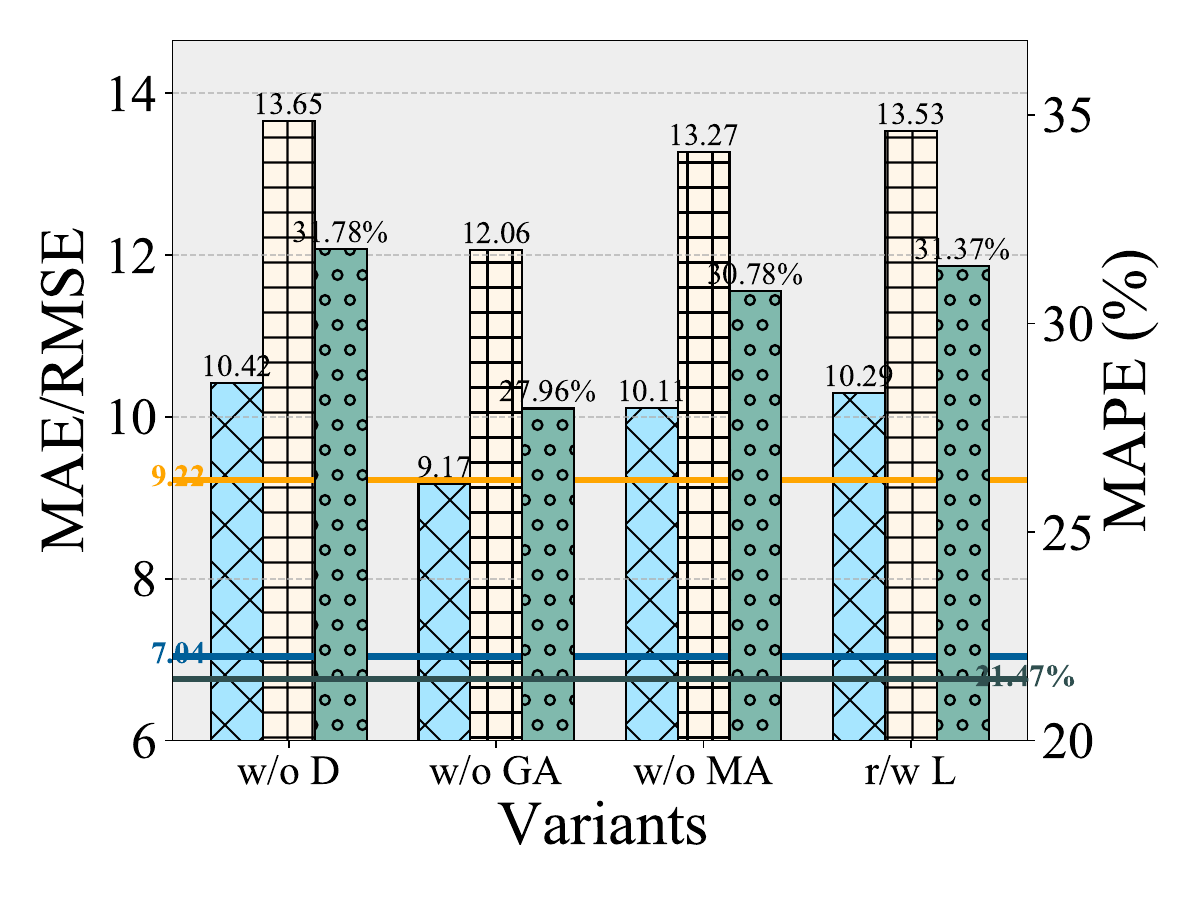}
}%
\subfigure[PeMS Traffic Flow]{
\vspace{-0.35cm}
\centering
% \hspace{0.35cm}
\includegraphics[width=0.45\linewidth]{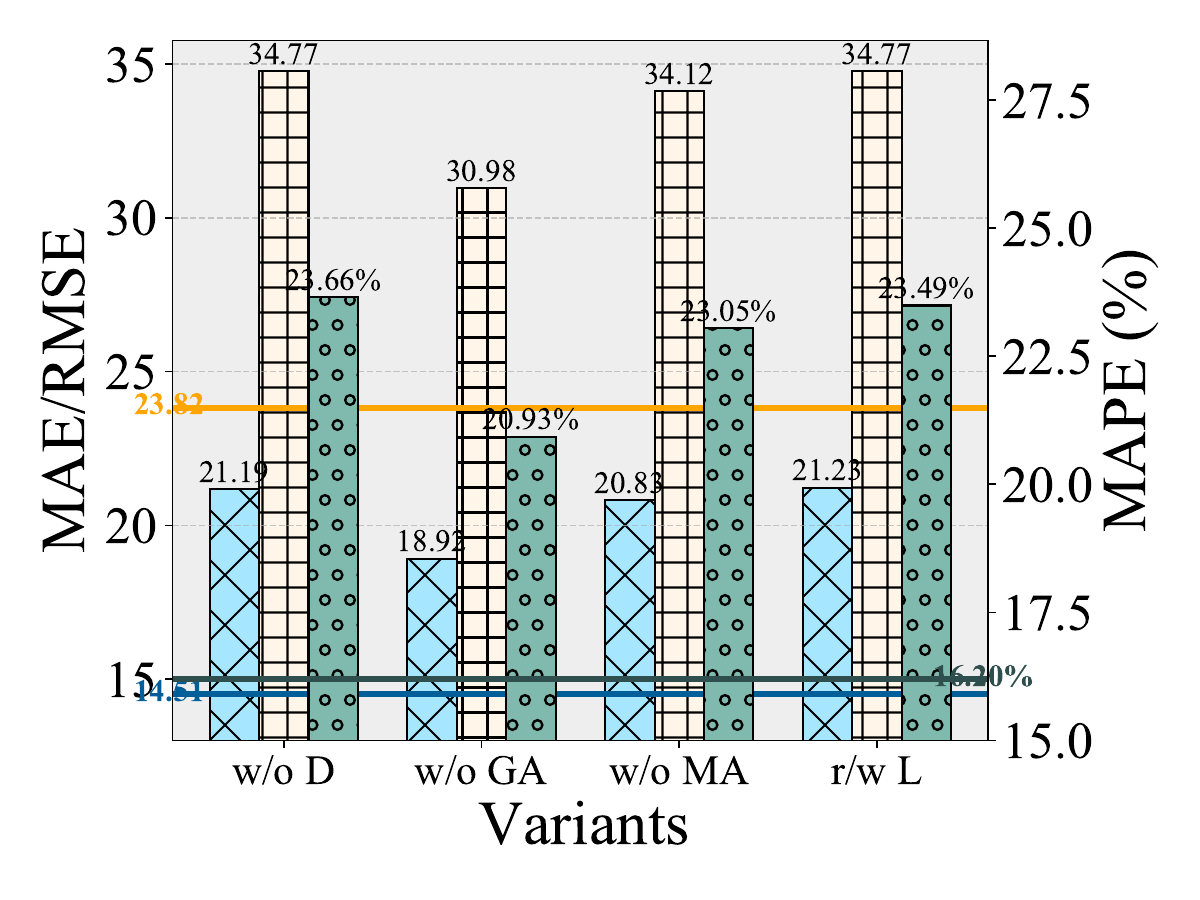}
}%
}
\vfill
\vspace{-0.35cm}
\subfigure[PeMS\_F Traffic Speed]{
\centering
\includegraphics[width=0.45\linewidth]{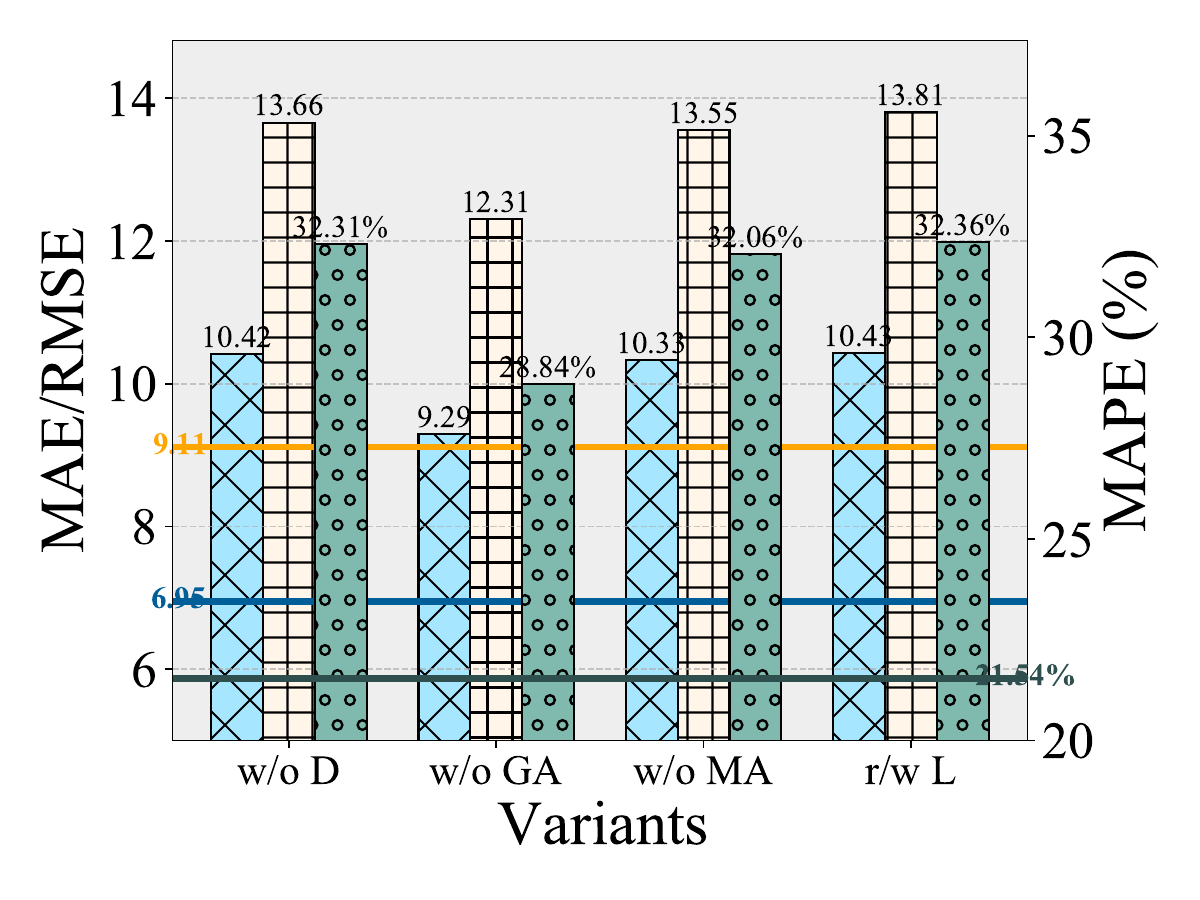}
}%
\subfigure[PeMS\_F Traffic Flow]{
\centering
\includegraphics[width=0.45\linewidth]{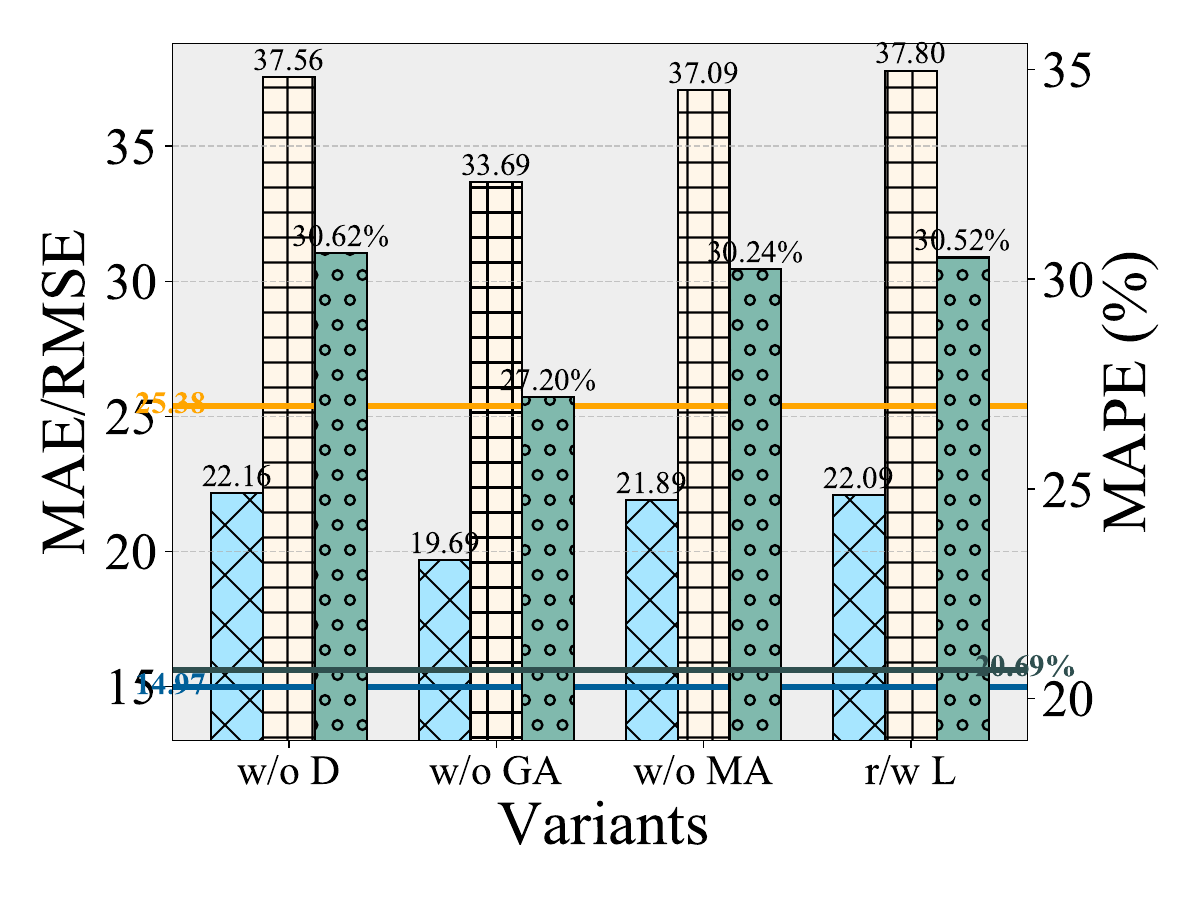}
}%
\centering
\vspace{-0.15cm}
\caption{Ablation results of RoadDiff components}
\label{fig: exp_ablation}
\end{figure}
\subsection{Ablation Study \label{sec: ablation}}
To evaluate the contribution of different components to the performance of the RoadDiff model, we conducted ablation experiments, including the removal of the Lane Diffusion Module (w/o D), the elimination of the Graph Attention (w/o GA) and MLP Attention mechanisms (w/o MA) within the Road-Lane Correlation Autoencoder-Decoder, and replacing the entire Road-Lane Correlation Autoencoder-Decoder module with a linear layer(r/w L). Since the Lane Diffusion Module requires input and output tensors to have the same shape, it could not be directly removed.  \textbf{Figure~\ref{fig: exp_ablation}} illustrates the performance comparison of RoadDiff and its variants on traffic flow and speed inference tasks using the PeMS and PeMS\_F datasets, with horizontal lines indicating RoadDiff's performance. The results demonstrate that removing any mechanism or module leads to a significant increase in inference errors for both traffic flow and speed. Notably, removing the Lane Diffusion Module resulted in the highest prediction errors across most datasets, further validating the critical role of the Lane Diffusion Module in constraint modeling and uncertainty handling.

\subsection{Study on Diffusion Steps}
\begin{table}[htbp]
  \centering
  \vspace{-0.25cm}
  \caption{Influence of Diffusion Steps}
  \vspace{-0.25cm}
      \resizebox{\linewidth}{!}{
    \begin{tabular}{cccccc|cccc}
    \toprule
    \multirow{2}[4]{*}{Datasets} & Traffic States & \multicolumn{4}{c}{Taffic Speed} & \multicolumn{4}{c}{Traffic Flow} \\
\cmidrule{2-10}        & \# Step & 5   & 10  & 20  & 30  & 5   & 10  & 20  & 30 \\
    \midrule
    \multirow{3}[2]{*}{PeMS} & MAE & 8.62  & 7.04  & 7.19  & 7.87  & 17.41  & 14.51  & 14.40  & 14.54  \\
        & RMSE & 11.29  & 9.22  & 11.50  & 10.30  & 28.58  & 23.82  & 23.64  & 23.88  \\
        & MAPE & 26.30\% & 21.47\% & 21.91\% & 23.99\% & 19.44\% & 16.20\% & 16.07\% & 16.24\% \\
    \midrule
    \multirow{3}[2]{*}{HuaNan} & MAE & 3.82  & 3.12  & 2.95  & 3.49  & 5.80  & 4.83  & 4.79  & 4.84  \\
        & RMSE & 5.40  & 4.41  & 4.29  & 4.93  & 9.42  & 7.85  & 7.79  & 7.87  \\
        & MAPE & 12.54\% & 10.24\% & 9.68\% & 11.44\% & 86.36\% & 71.97\% & 71.41\% & 72.16\% \\
    \midrule
    \multirow{3}[2]{*}{PeMS\_F} & MAE & 8.51  & 6.95  & 7.09  & 7.76  & 17.96  & 14.97  & 14.85  & 14.56  \\
        & RMSE & 11.16  & 9.11  & 11.36  & 10.18  & 30.46  & 25.38  & 25.18  & 24.69  \\
        & MAPE & 26.39\% & 21.54\% & 21.98\% & 24.06\% & 24.83\% & 20.69\% & 20.53\% & 20.13\% \\
    \bottomrule
    \end{tabular}%
    }
  \label{tab: diffusion}%
  % \vspace{-0.35cm}
\end{table}%
The diffusion step parameter determines the number of noise addition or denoising iterations and the extent of noise/denoising applied in each step. To explore the impact of diffusion steps on the inference accuracy of the RoadDiff framework, we compared the inference errors for 5, 10, 20, and 30 diffusion steps on the traffic speed and flow data of the PeMS, HuaNan, and PeMS\_F datasets. The results in \textbf{Table \ref{tab: diffusion}} show that diffusion steps have a significant impact on traffic speed inference. When the step count is 5, the performance on all three datasets is the worst, while the best performance occurs at 10 or 20 steps. This indicates that too few steps might prevent the model from adequately recovering data details, requiring each step to reduce more noise, and resulting in poorer data generation quality. The performance at 30 steps is not as good as at 10 or 20 steps, likely due to overfitting caused by too many steps for the limited information in the data, leading to poorer performance on the test set. For traffic flow inference, the disadvantage of fewer steps is even more pronounced, with the best performance occurring at 10, 20, and 30 steps. This might be due to the greater variability and differences between road flow data and lane flow data, where more steps help in better fitting. Overall, setting an appropriate number of diffusion steps significantly impacts the model's inference capability and should be adjusted based on the characteristics of the dataset.

\section{Related Work}
\subsection{Graph-based Generation}

Graphs encode relational information across many fields. Graph generation aims to create new graphs from distributions similar to observed graphs \cite{zhu2203survey}. Current research focuses on molecular design, protein design, and program synthesis.

In molecular generation, methods must generate syntactically valid molecules, and ensure the generated molecules have specific properties. GraphNVP \cite{madhawa2019graphnvp} and GRF \cite{honda2019graph} first introduced normalization flow-based models for molecular graph generation. MoFlow \cite{zang2020moflow} improved this with a valence correction step. Proteins are represented by pairwise contact graphs based on 3D structural data. CO-VAE \cite{guo2021generating} designed a VAE-based model to generate new protein contact graphs and decode their 3D structure. RefineGNN \cite{jiniterative} introduced an iteratively improved GNN model to design the sequence and structure of antibody complementarity-determining regions (CDR). Program synthesis aims to generate programs from natural language descriptions and input-output samples. Researchers proposed program graphs encompassing syntax and semantic knowledge. ExprGen \cite{brockschmidt2018generative} enhanced partially generated ASTs with additional syntax and semantic connections, while SD-VAE \cite{dai2018syntax} used attribute grammars to improve the semantic validity of generated program graphs.

Despite advancements, these methods are too domain-specific for spatio-temporal graph node generation. However, spatio-temporal graph data are critical in traffic, social networks, air monitoring, and other fields, making effective solutions for spatio-temporal graph node generation a key research topic.

\subsection{Fine-grained Urban Flow Inference}
Fine-grained urban flow inference aims to obtain detailed traffic flow data using fewer sensors, reducing manpower and energy costs for smart cities. UrbanFM \cite{liang2019urbanfm} was the first to address the fine-grained urban flow inference (FUFI) problem, proposing a distributed upsampling module and an external factor fusion subnet. Subsequent work improved UrbanFM in spatial constraints, external factors, and storage costs. FODE \cite{zhou2020enhancing} and UrbanODE \cite{zhou2021inferring}, based on neural ordinary differential equations (ODEs), tackle numerical instability in FUFI through affine coupling layers and pyramid attention networks. MT-CSR \cite{li2022fine} addresses FUFI in incomplete urban flow maps. DeepLGR \cite{liang2021revisiting} reconsiders CNN limitations, learning global spatial dependencies and local feature representations. UrbanPy \cite{ouyang2020fine} extends UrbanFM with a pyramid-based cascade strategy, proposal and correction components, and a new distribution loss. CUFAR \cite{yu2023overcoming} introduces adaptive knowledge replay (AKR) to overcome "catastrophic forgetting" in FUFI tasks, achieving accurate inference.

Similarly, fine-grained road traffic inference aims to avoid the inefficiencies of obtaining lane data from video sensors or adding extra lane sensors. Compared to grid-based urban flow data, the FRTI task faces more constraints and a more flexible road network structure.

\section{Conclusions and Future Work}
In this paper, we propose the spatio-temporal graph node generation problem and its corresponding fine-grained road traffic inference task, highlighting challenges like limited information, variable road topology, and theoretical constraints between traffic states. To address these challenges, we present a two-stage framework—RoadDiff—based on an autoencoder structure and a denoising diffusion generation module. RoadDiff models the road network via a graph structure, capturing spatio-temporal dependencies between roads and lanes to generate initial lane traffic states. The lane diffusion module further models real-world uncertainties and approximates real lane data distributions, generating accurate lane information. This method produces granular lane-level spatio-temporal graphs from small-scale road graphs, offering energy-efficient and precise data support.

Future work will enhance the diffusion model's efficiency and optimize computational performance for larger datasets. Additionally, we will explore and validate the graph diffusion model's potential in other spatio-temporal graph generation problems.

\section*{ACKNOWLEDGMENTS}
We thank the anonymous reviewers for their valuable suggestions on improving our work and guiding future research. The research work described in this paper was supported by Hong Kong Research Grants Council (grant\# 16202722, T22-607/24-N, T43-513/23N-1). This work was conducted at the BDKE Lab of the School of Computer Science at Fudan University and partially conducted in JC STEM Lab of Data Science Foundations funded by The Hong Kong Jockey Club Charities Trust.
\clearpage
%%
%% The next two lines define the bibliography style to be used, and
%% the bibliography file.
\bibliographystyle{ACM-Reference-Format}
\balance
\bibliography{reference}
\clearpage
%%
%% If your work has an appendix, this is the place to put it.
\appendix
\section*{Appendix}
\section{Constraint Analysis}

In this section, we provide a detailed theoretical analysis of the two critical constraints incorporated into the RoadDiff framework: the Traffic Speed Constraint and the Traffic Flow Constraint. These constraints are based on fundamental traffic laws—namely, the Traffic Flow Conservation Law and the Traffic Speed Consistency Law—which ensure the physical coherence and accuracy of traffic state inference.

\subsection{Traffic Flow Constraint}

\textbf{Law 1 (Traffic Flow Conservation Law):} \textit{The Traffic Flow Conservation Law states that in a traffic system, unless there is external interference, the number of vehicles passing through a specific section remains constant over a given period of time.} That is, vehicle numbers cannot spontaneously increase or disappear. For a road segment \( r_i \), the input and output traffic flow at time \( t \) must remain balanced, ensuring that the total vehicle count remains unchanged \cite{daganzo1997fundamentals}. This principle ensures that for a multi-lane road, the total traffic flow of a road segment equals the sum of the traffic flows of its individual lanes.

For a road segment \( r_i \) with \( J_i \) lanes at time \( t \), the flow conservation can be expressed as:
\begin{equation}
    Q^{r_i}_t = \sum_{j=1}^{J_i} Q^{l_{i,j}}_t,
\end{equation}
where \( Q^{r_i}_t \) represents the total traffic flow of the road segment \( r_i \), and \( Q^{l_{i,j}}_t \) represents the traffic flow of lane \( l_{i,j} \).

Traffic flow (\( Q \)) is fundamentally determined by the relationship:
\begin{equation}
    Q = K \cdot V,
\end{equation}
where \( K \) denotes traffic density (vehicles per unit length) and \( V \) denotes traffic speed (length per unit time). Applying this relationship to a multi-lane road, the total flow can be expressed as:
\begin{equation}
    Q^{r_i}_t = \sum_{j=1}^{J_i} K^{l_{i,j}}_t \cdot V^{l_{i,j}}_t.
\end{equation}

The Traffic Flow Constraint thus ensures that the RoadDiff model adheres to the principle of vehicle conservation when inferring lane-level traffic information from road-level data:
\begin{equation}
    x_t^{r_i} = \sum_{j=1}^{J_i} x_t^{l_{i,j}}.
\end{equation}

By enforcing this constraint, the model captures the intrinsic relationship between road and lane traffic flows, ensuring that vehicle dynamics are preserved across scales. This constraint is particularly crucial for traffic flow inference in high-density or highly dynamic scenarios, where deviations from conservation laws could lead to unrealistic or inaccurate predictions. Moreover, incorporating this constraint into the training loss not only enhances the physical plausibility of the predicted values but also helps regularize the learning process by discouraging solutions that violate basic traffic flow consistency.

\subsection{Traffic Speed Constraint}

\textbf{Law 2 (Traffic Speed Consistency Law):} \textit{The Traffic Speed Consistency Law states that, under specific equilibrium traffic conditions, the average speed of lanes on the same road should equal the overall road speed} \cite{greenshields1935study}. This principle ensures coherence between the macroscopic and microscopic views of traffic speed.

For a road segment \( r_i \) with \( J_i \) lanes at time \( t \), the average speed relationship can be expressed as:
\begin{equation}
    V^{r_i}_t = \frac{1}{J_i} \sum_{j=1}^{J_i} V^{l_{i,j}}_t,
\end{equation}
where \( V^{r_i}_t \) represents the speed of the road segment \( r_i \), and \( V^{l_{i,j}}_t \) represents the speed of lane \( l_{i,j} \).

In equilibrium traffic conditions, the density (\( K \)) of the road segment can be considered relatively uniform across lanes. Thus, the relationship \( Q = K \cdot V \) implies that speed consistency is critical for ensuring coherence in traffic state inference:
\begin{equation}
    x_t^{r_i} = \frac{1}{J_i} \sum_{j=1}^{J_i} x_t^{l_{i,j}}.
\end{equation}

The Traffic Speed Constraint ensures that the RoadDiff model maintains this coherence by aligning the road speed with the average speed of its constituent lanes. This alignment is essential for tasks such as lane-specific routing, where accurate speed estimation directly impacts the quality of routing decisions.

\section{Evaluation Metric Formulas \label{app:metrics}}
We used three of the most common evaluation metrics in the field of traffic:

\noindent\textbf{Mean Absolute Error (MAE)}: Measures the average absolute difference between the predicted values and the actual values.
\begin{equation}
MAE = \frac{1}{T}\sum^{T}_{t = 1}\frac{1}{N}\sum^{I}_{i = 1}\sum^{J_i}_{j = 1}\Big|x^{l_{i,j}}_t - \widehat{x}^{l_{i,j}}_t \Big|
\end{equation}
\noindent\textbf{Root Mean Squared Error (RMSE)}: Reflects the square root of the average squared differences between predicted values and actual values, emphasizing the impact of larger errors.
\begin{equation}
RMSE = \sqrt{\frac{1}{T}\sum^{T}_{t = 1}\frac{1}{N}\sum^{I}_{i = 1}\sum^{J_i}_{j = 1}\Big(x^{l_{i,j}}_t - \widehat{x}^{l_{i,j}}_t \Big)^2}
\end{equation}
\noindent\textbf{Mean Absolute Percentage Error (MAPE)}: Measures the percentage difference between the predicted and actual values, reflecting the size of the relative error.
\begin{equation}
MAPE = \frac{1}{T}\sum^{T}_{t = 1}\frac{1}{N}\sum^{I}_{i = 1}\sum^{J_i}_{j = 1}\Big|\frac{x^{l_{i,j}}_t - \widehat{x}^{l_{i,j}}_t}{x^{l_{i,j}}_t} \Big|
\end{equation}
\section{Supplemental Experiments \label{app:exp}}
We present additional experimental results and analysis on the HuaNan dataset, which were omitted from the main text due to space limitations. These include the main experimental results, time window size study, and ablation analysis.

\subsection{Main Experiment}
\begin{table*}[h]
  \centering
  % \vspace{-0.45cm}
  \caption{Comparison on HuaNan Datasets}
  \vspace{-0.25cm}
  \resizebox{0.9\linewidth}{!}{
    \begin{tabular}{lcccccccccc|ccccccccc}
    \toprule
    \multicolumn{1}{c}{\multirow{2}[4]{*}{Model}} & \multicolumn{2}{c}{Traffic Type} &     &     & \multicolumn{3}{c}{ Traffic Speed} &     &     & \multicolumn{1}{r}{} & \multicolumn{9}{c}{ Traffic Flow} \\
\cmidrule{2-20}        & \multicolumn{2}{c}{Time Windows} & 1   &     & \multicolumn{3}{c}{3} & \multicolumn{3}{c|}{6} & \multicolumn{3}{c}{1} & \multicolumn{3}{c}{3} & \multicolumn{3}{c}{6} \\
\cmidrule{2-20}    \multicolumn{1}{c}{Types} & \multicolumn{1}{l}{Metrics} & MAE & RMSE & MAPE & MAE & RMSE & MAPE & MAE & RMSE & MAPE & MAE & RMSE & MAPE & MAE & RMSE & MAPE & MAE & RMSE & MAPE \\
    \midrule
    \multicolumn{1}{c}{\textbf{Basic}} & Physics & $\underline{5.91}$ & $\underline{7.99}$ & $\underline{17.73\%}$ & $\underline{5.92}$ & $\underline{8.02}$ & $\underline{17.72\%}$ & $\underline{5.91}$ & $\underline{7.92}$ & $\underline{17.70\%}$ & $\underline{7.08}$ & $\underline{11.78}$ & $\underline{133.40\%}$ & $\underline{7.11}$ & $\underline{10.75}$ & $\underline{134.50\%}$ & $\underline{7.78}$ & $\underline{12.35}$ & $\underline{133.34\%}$ \\
    \midrule
    \multicolumn{1}{c}{\multirow{5}[2]{*}{\begin{sideways}\textbf{FUFI}\end{sideways}}} & UrbanFM & 6.04  & 7.15  & 19.17\% & 6.02  & 7.13  & 19.10\% & 6.04  & 7.16  & 19.51\% & 5.44  & 11.43  & 88.59\% & 5.44  & 11.43  & 88.59\% & 5.44  & 11.43  & 88.58\% \\
        & FODE & 6.14  & 7.49  & 23.30\% & 6.12  & 7.47  & 23.21\% & 6.13  & 7.54  & 23.91\% & 5.53  & 11.98  & 107.68\% & 5.53  & 11.97  & 107.68\% & 5.53  & 11.97  & 107.67\% \\
        & UrbanPy & 5.99  & 7.08  & 19.02\% & 5.97  & 7.07  & 18.95\% & 5.98  & 7.07  & 18.97\% & 5.40  & 11.33  & 87.91\% & 5.40  & 11.32  & 87.91\% & 5.40  & 11.32  & 87.90\% \\
        & DeepLGR & 6.38  & 7.35  & 19.60\% & 6.36  & 7.34  & 19.52\% & 6.36  & 7.38  & 19.56\% & 5.75  & 11.76  & 90.58\% & 5.75  & 11.75  & 90.58\% & 5.75  & 11.76  & 90.57\% \\
        & CUFAR & $\underline{5.53}$ & $\underline{6.75}$ & $\underline{17.51\%}$ & $\underline{5.51}$ & $\underline{6.74}$ & $\underline{17.44\%}$ & $\underline{5.50}$ & $\underline{6.76}$ & $\underline{17.76\%}$ & $\underline{5.09}$ & $\underline{10.10}$ & $\underline{80.12\%}$ & $\underline{5.08}$ & $\underline{10.80}$ & $\underline{80.92\%}$ & $\underline{5.08}$ & $\underline{10.99}$ & $\underline{82.91\%}$ \\
    \midrule
    \multicolumn{1}{c}{\multirow{4}[2]{*}{\begin{sideways}\textbf{ED}\end{sideways}}} & DCRNN & 8.33  & 10.02  & 25.74\% & 8.32  & 10.02  & 25.69\% & 8.33  & 10.02  & 25.78\% & 8.46  & 15.88  & 149.97\% & 8.46  & 15.90  & 149.96\% & 8.45  & 15.87  & 149.86\% \\
        & GMAN & 7.55  & 8.75  & 23.32\% & 7.54  & 8.75  & 23.28\% & 7.56  & 8.75  & 23.35\% & 7.67  & 13.90  & 135.86\% & 7.66  & 13.80  & 135.81\% & 7.67  & 13.88  & 135.84\% \\
        & Bi-STAT & $\underline{7.01}$ & $\underline{8.53}$ & 21.04\% & $\underline{7.00}$ & $\underline{8.53}$ & 21.01\% & $\underline{7.01}$ & $\underline{8.53}$ & 21.08\% & $\underline{7.11}$ & $\underline{13.53}$ & 122.49\% & $\underline{7.12}$ & $\underline{13.54}$ & 122.60\% & $\underline{7.12}$ & $\underline{13.56}$ & 122.61\% \\
        & MegaCRN & 7.74  & 9.81  & $\underline{18.83\%}$ & 7.73  & 9.81  & $\underline{18.80\%}$ & 7.75  & 9.81  & $\underline{18.86\%}$ & 7.96  & 15.76  & $\underline{111.34\%}$ & 7.94  & 15.74  & $\underline{111.80\%}$ & 7.95  & 15.75  & $\underline{113.72\%}$ \\
    \midrule
    \multicolumn{1}{c}{\multirow{7}[2]{*}{\begin{sideways}\textbf{STG + Linear}\end{sideways}}} & STGCN & 8.06  & 9.81  & 24.62\% & 8.06  & 9.81  & 24.57\% & 8.07  & 9.81  & 24.66\% & 7.86  & 14.90  & 137.74\% & 7.93  & 15.08  & 138.71\% & 8.45  & 18.01  & 110.77\% \\
        & MTGNN & $\underline{5.41}$ & $\underline{7.27}$ & $\underline{15.36\%}$ & $\underline{5.41}$ & $\underline{7.28}$ & $\underline{15.34\%}$ & 5.42  & 7.28  & $\underline{15.39\%}$ & 5.34  & 11.25  & 86.89\% & 5.33  & $\underline{11.20}$ & $\underline{86.82\%}$ & $\underline{5.33}$ & $\underline{11.20}$ & $\underline{86.82\%}$ \\
        & ASTGCN & 7.37  & 8.85  & 21.08\% & 7.30  & 8.74  & 20.89\% & 6.86  & 7.32  & 26.26\% & 7.19  & 13.44  & 117.97\% & 7.19  & 13.44  & 117.95\% & 7.19  & 13.44  & 117.97\% \\
        & GraphWaveNet & 6.17  & 8.31  & 16.21\% & 6.11  & 8.21  & 16.06\% & 5.75  & 6.87  & 20.18\% & 6.08  & 12.79  & 91.59\% & 6.03  & 12.63  & 90.92\% & 5.66  & 10.58  & 113.87\% \\
        & STSGCN & 6.39  & 8.18  & 16.43\% & 6.38  & 8.18  & 16.40\% & 6.39  & 8.19  & 16.46\% & 6.23  & 12.43  & 91.93\% & 6.28  & 12.58  & 92.59\% & 6.69  & 15.02  & 93.93\% \\
        & AGCRN & 5.47  & 7.37  & 15.52\% & 5.42  & 7.28  & 15.38\% & $\underline{5.10}$ & $\underline{6.10}$ & 19.33\% & $\underline{5.26}$ & $\underline{11.20}$ & $\underline{86.83\%}$ & $\underline{5.31}$ & 11.44  & 90.07\% & 5.34  & 11.22  & 86.92\% \\
        & STGODE & 6.16  & 7.95  & 15.77\% & 6.10  & 7.85  & 15.63\% & 5.74  & 6.58  & 19.64\% & 6.01  & 12.07  & 88.22\% & 6.01  & 12.07  & 88.21\% & 6.01  & 12.07  & 88.22\% \\
    \midrule
    \textbf{Ours} & RoadDiff & \textit{\textbf{3.15 }} & \textit{\textbf{4.46 }} & \textit{\textbf{10.32\%}} & \textit{\textbf{3.12 }} & \textit{\textbf{4.41 }} & \textit{\textbf{10.24\%}} & \textit{\textbf{3.11 }} & \textit{\textbf{4.40 }} & \textit{\textbf{10.21\%}} & \textit{\textbf{4.88 }} & \textit{\textbf{7.94 }} & \textit{\textbf{72.55\%}} & \textit{\textbf{4.83 }} & \textit{\textbf{7.85 }} & \textit{\textbf{71.97\%}} & \textit{\textbf{4.88 }} & \textit{\textbf{7.94 }} & \textit{\textbf{72.55\%}} \\
    \bottomrule
    \end{tabular}%
    }
  \label{tab: mian_HuaNan}%
   % \vspace{-0.25cm}
\end{table*}%
\textbf{Table \ref{tab: mian_HuaNan}}  presents the performance of various models on the HuaNan Road-Lane dataset for traffic speed and flow inference tasks. In the traffic speed dataset, FUFI models demonstrated overall superior performance, significantly outperforming other baseline categories. Conversely, ED-based models performed the worst, likely due to their inability to model multi-granularity relationships effectively, highlighting the unique challenges of the FRTI problem. RoadDiff outperformed all baseline models in inference accuracy, further showcasing its exceptional performance across various datasets. Meanwhile, other baseline models such as CUFAR, MTGNN, and AGCRN also achieved competitive results. For the traffic flow inference task, RoadDiff again delivered the best performance, followed by the CUFAR model. This suggests that deep learning-based urban traffic flow inference models can be effectively adapted to lane-level flow inference tasks in scenarios with regular lane counts, offering robust methodological support for addressing this challenge.

\begin{figure} [htbp]
\centerline{
\vspace{-0.45cm}
\subfigure[HuaNan Speed Dataset]{
\centering
\includegraphics[width=0.45\linewidth]{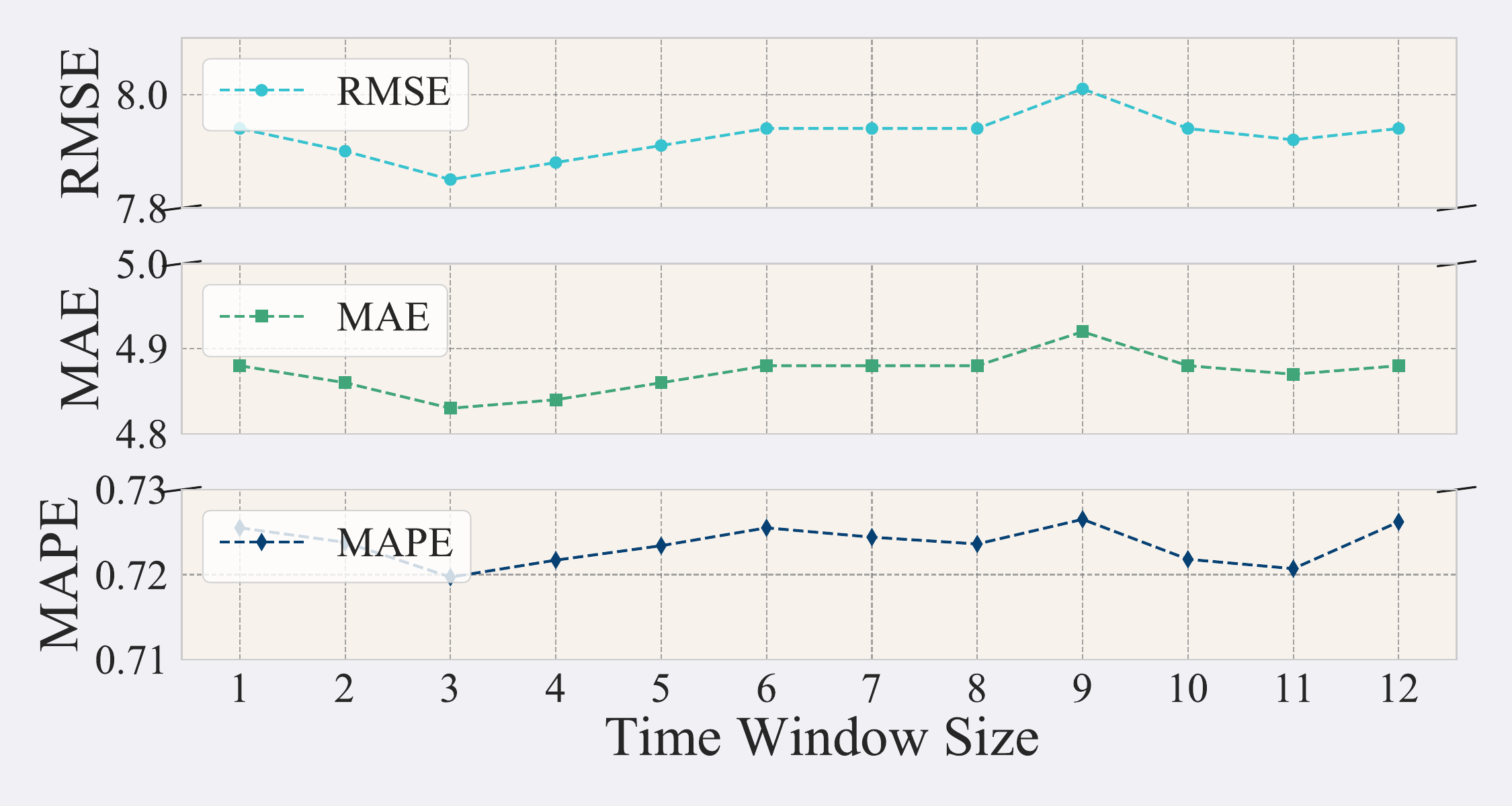}
}%
\subfigure[HuaNan Flow Dataset]{
\centering
\includegraphics[width=0.45\linewidth]{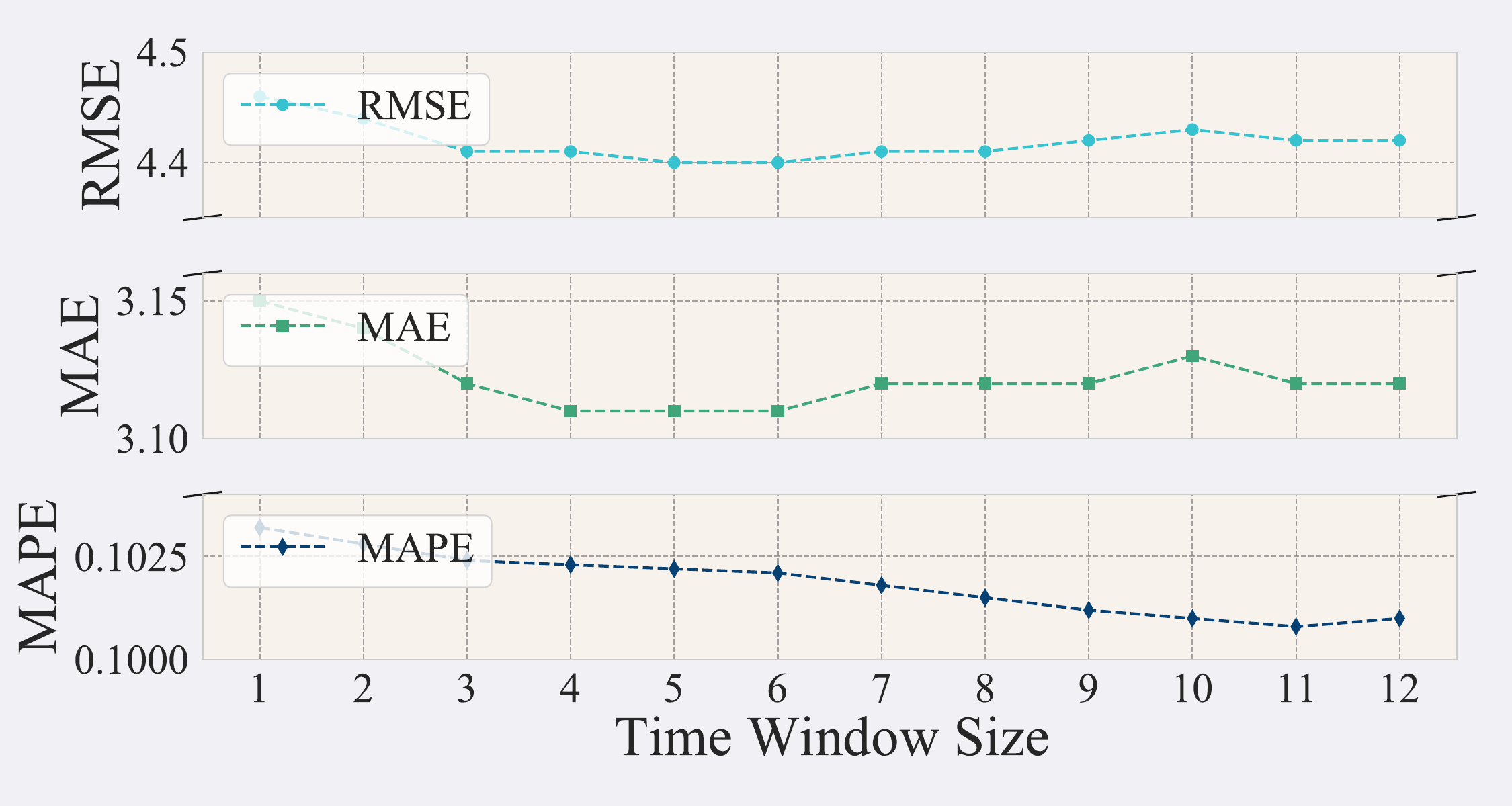}
}%
}
\centering
% \vspace{-0.15cm}
\caption{Influence of Time Window Size on HuaNan Datasets.}
\label{fig: appendix_time_window}
\vspace{-0.55cm}
\end{figure}

\subsection{Study on Time Window Size}
\textbf{Figure \ref{fig: appendix_time_window}} shows the performance of RoadDiff on the HuaNan dataset with different time window sizes. For traffic speed, similar to the freeway datasets PeMS and PeMS\_F, despite slight fluctuations, RoadDiff can accurately infer traffic speed over longer time windows within the 1-12 window scale. In terms of traffic flow inference, fewer time windows result in higher inference errors, indicating that traffic flow inference relies more on the spatio-temporal relationship information of roads and lanes across both types of datasets.

\subsection{Ablation Study}
\begin{figure} [htbp]
% \vspace{-0.30cm}
\includegraphics[width=0.99\linewidth]{fig/exp/detailed_legend.pdf} 
 \vspace{-0.15cm} 
\newline
\centerline{
\subfigure[HuaNan Speed Dataset]{
\centering
\includegraphics[width=0.45\linewidth]{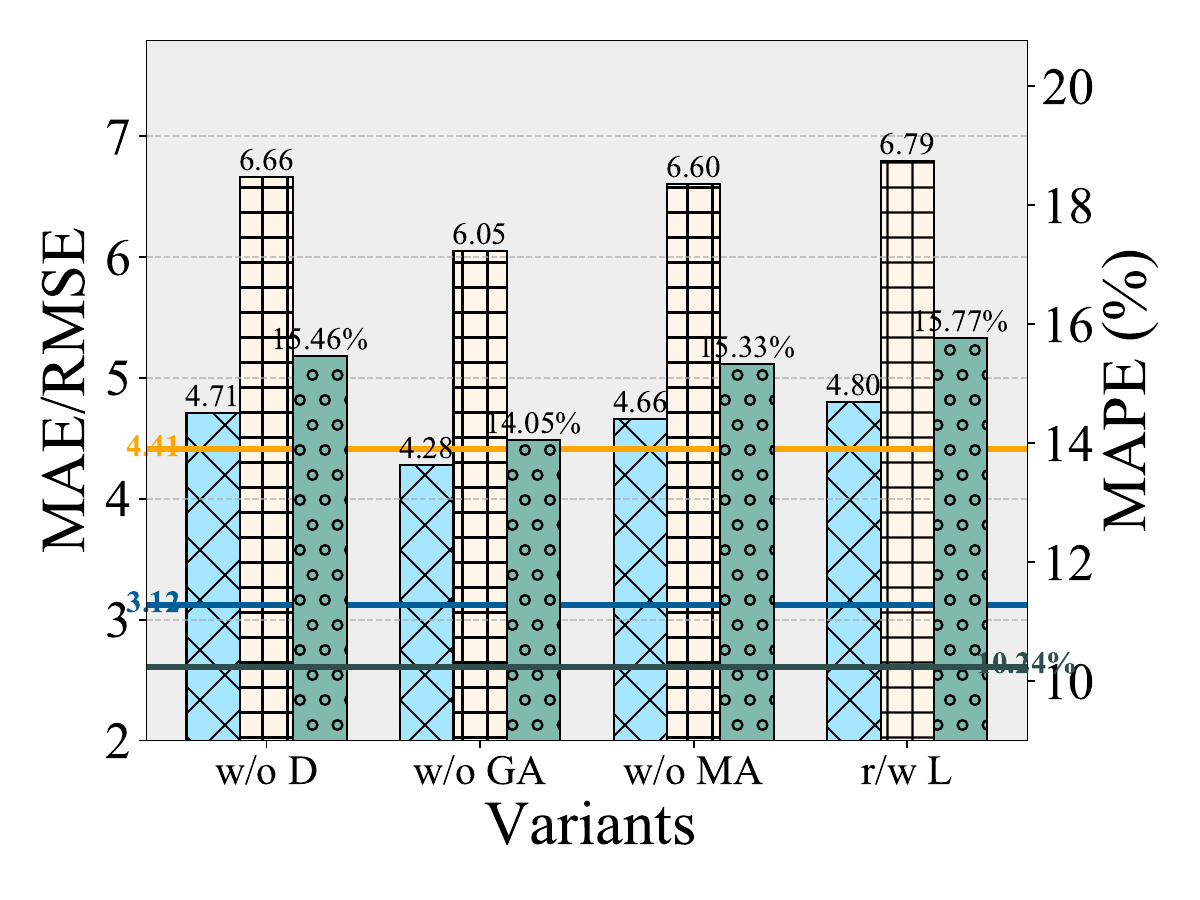}
}%
\subfigure[HuaNan Flow Dataset]{
\centering
\includegraphics[width=0.45\linewidth]{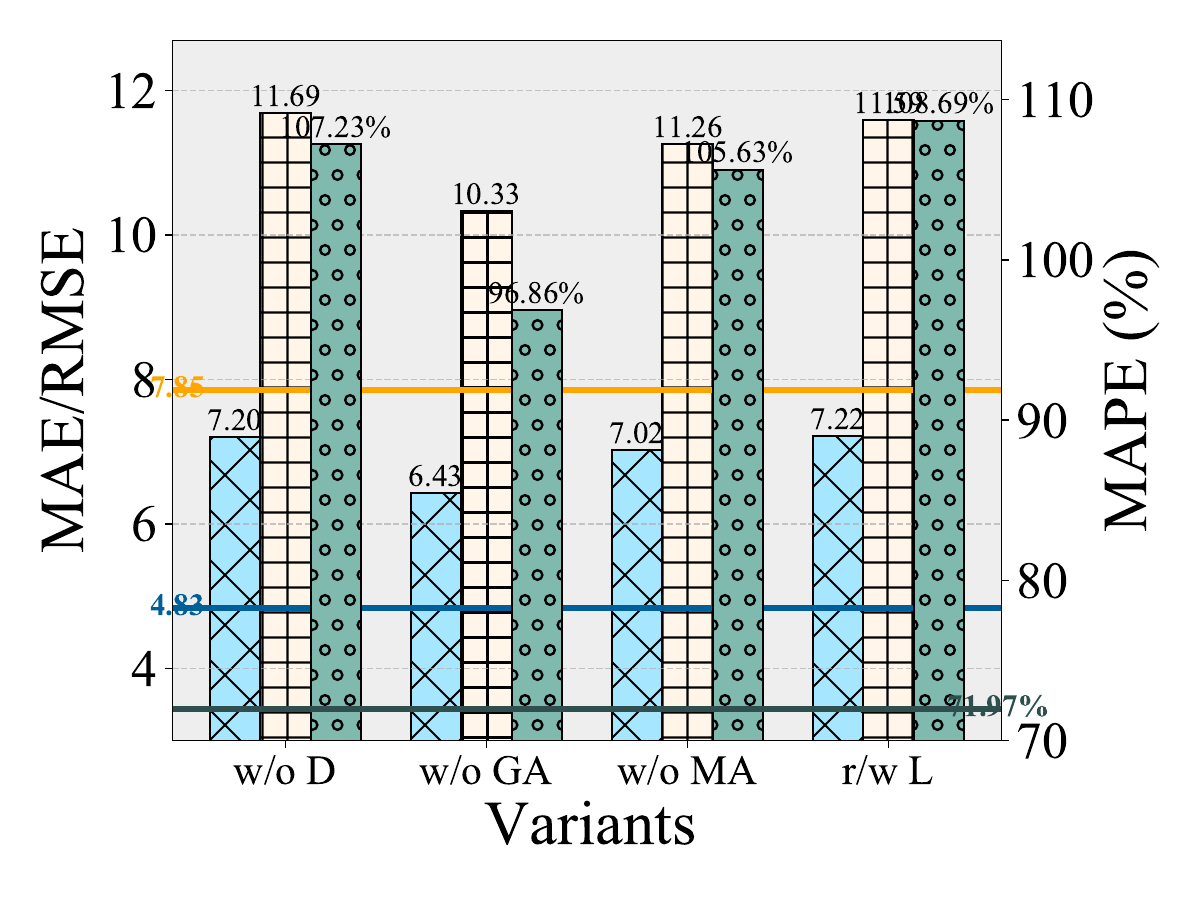}
}%
}
\centering
\vspace{-0.25cm}
\caption{Ablation on HuaNan Datasets.}
\label{fig: appendix_ablation}
\end{figure}

\textbf{Figure \ref{fig: appendix_ablation}} compares the performance of RoadDiff variants with individual components removed on the HuaNan dataset for traffic flow and speed data. Unlike the results on the PeMS and PeMS\_F datasets, replacing the entire Road-Lane Correlation Autoencoder-Decoder module with a linear layer resulted in a larger error. This indicates that robust and effective road-lane correlation modeling provides better initial information to support the Lane Diffusion module. Although the Road-Lane Correlation Autoencoder-Decoder module can independently address the FRTI problem, its standalone use leads to significantly higher errors. The experimental results confirm the critical importance of the synergy between the two modules in achieving optimal inference performance.

\end{document}